\documentclass{article}

\usepackage[numbers,sort,compress]{natbib}

\usepackage[preprint]{neurips_2020}

\usepackage[utf8]{inputenc} 
\usepackage[T1]{fontenc}    
\usepackage{hyperref}       
\usepackage{url}            
\usepackage{booktabs}       
\usepackage{amsfonts}       
\usepackage{nicefrac}       
\usepackage{microtype}      
\usepackage{graphicx}

\usepackage{subcaption}
\usepackage{todonotes}
\usepackage{array}
\usepackage{amsfonts, amsmath, amssymb}
\usepackage[ruled,linesnumbered]{algorithm2e}
\include{pythonlisting}

\title{Contrastive Weight Regularization for Large Minibatch SGD}
\author{%
  Qiwei Yuan$^1$, Weizhe Hua$^2$, Yi Zhou$^1$, Cunxi Yu$^1$\\
  University of Utah$^1$\\
  Cornell University$^2$ \\
  \texttt{cunxi.yu@utah.edu} 
}

\begin{document}

\maketitle

\begin{abstract}


The minibatch stochastic gradient descent method (SGD) is widely applied in deep learning due to its efficiency and scalability that enable training deep networks with a large volume of data. Particularly in the distributed setting, SGD is usually applied with a large batch size. However, as opposed to small-batch SGD, neural network models trained with large-batch SGD can hardly generalize well, i.e., the validation accuracy is low. In this work, we introduce a novel regularization technique, namely distinctive regularization (DReg), which replicates a certain layer of the deep network and encourages the parameters of both layers to be diverse. The DReg technique introduces very little computation overhead. Moreover, we empirically show that optimizing the neural network with DReg using large-batch SGD achieves a significant boost in the convergence and an improved generalization performance. We also demonstrate that DReg can boost the convergence of large-batch SGD with momentum. We believe that DReg can be used as a simple regularization trick to accelerate large-batch training in deep learning.
\end{abstract}

\section{Introduction}


Minibatch SGD is the dominant algorithm for optimizing deep neural networks (DNNs) derived from SGD. While large datasets and complex DNNs benefit the performance of deep learning tasks, effective training of complex models with large datasets is very challenging due to the significant growth in the batch size \cite{lin2017deep,akiba2017extremely,you2017imagenet,you2019reducing}. While large batch size increases the resource utilization of modern distributed computing systems, algorithms with large minibatches face generalization difficulties \cite{goyal2017accurate,wu2018group,hoffer2017train,smith2017don,peng2018megdet}, i.e., the \textit{generalization gap} \cite{keskar2016large}. In other words, increasing batch size often leads to a significant loss in test accuracy, due to the lack of generalization caused by the fact of converging to a sharp minimizer using large minibatch SGD \cite{keskar2016large}. There has been a significant amount of effort in addressing the generalization gap. For example, controlling the learning rate has been extensively studied, such as linear scaling learning rate \cite{krizhevsky2014one}, sharp decaying \cite{zagoruyko2016wide}, dynamic scaling \cite{zhao2019adaptive}, etc. Moreover, recent progress demonstrates that leveraging gradient noise for large minibatch training can be beneficial, especially in non-convex optimization \cite{zhou2018,welling2011bayesian,smith2017bayesian}. However, we found limited literature in addressing the generalization gap using SGD regularization algorithms, which directly improves the gradient generalizability of each batch. More importantly, an orthogonal regularization algorithm can be combined together to close the generalization gap further.

\paragraph{Contributions} In this work, we propose a novel regularization algorithm for minibatch SGD, which can be combined with the aforementioned existing techniques to address the generalization gap issue. Our main contributions are summarized as follows: \textbf{(1)} We propose a novel lightweight regularization technique inspired by \cite{hadsell2006dimensionality}, namely \textit{distinctive regularization} (DReg), to boost minibatch SGD optimization on Convolution Neural Network (ConvNet). DReg increases the generalizability of the gradients of each batch compared to vanilla SGD. \textbf{(2)} We provide a comprehensive evaluation of DReg with a wide range of batch size (from $2^{10}$ to $2^{10} \cdot 30$) using various popular ConvNet models and datasets. \textbf{(3)} We empirically analyze the performance impacts of various DReg configurations and the effects of DReg on momentum in SGD.   

\section{Background}
\textbf{Minibatch SGD} Let $f(\cdot; W)$ be a DNN model parameterized by $W$, which is trained on a finite training set $X$ of $M$ samples. Train a network using SGD is to minimize the loss function $\mathcal{L}(W)$, which is the average of the loss $l$ of every training data in the training set. Instead of updating the parameters $W$ based on a single training example, minibatch SGD \cite{minibatchSGD} updates the $W$ using the gradients obtained from a minibatch, with $\eta$ as the stepsize (i.e., learning rate) and $t$ is the index of the training iteration (Equation \ref{eq:minibatch}).
\begin{equation}
    \label{eq:minibatch}
    \mathcal{L}(W) = \frac{1}{M}\sum_{x \in X}l(x, W);~~~~~~~~W_{(t+1)} = W_{(t)} - \eta \frac{1}{m}\sum_{x \in \mathcal{B}}\nabla l(x, W_{(t)}),
\end{equation}

Moreover, momentum SGD \cite{momentumSGD} is a widely-adopted method to accelerate the vanilla minibatch SGD in Equation~\ref{eq:minibatch}.
To update the parameters with less noisy gradients, a momentum is defined as a moving average of the minibatch gradients and used to update the parameters of the form:
\begin{equation}
\label{eq:momentum}
    v_{(t+1)} = \beta v_{(t)} + \frac{1}{m}\sum_{x \in \mathcal{B}}\nabla l(x, W_{(t)});~~~~~~~~
    W_{(t+1)} = W_{(t)} - \eta v_{(t+1)}
\end{equation}
Here, $\beta$ denotes the momentum decaying factor and $v$ is the update tensor.
The momentum SGD can be considered as the state-of-the-art optimization algorithm for DNNs.
The proposed distinctive regularization method can speedup the convergence and improve the generalization of DNNs trained using both the vanilla and momentum SGD.

\textbf{Large Minibatch SGD} Many recent works \cite{fb_largebatch, google_largebatch} propose to scale up the training of large DNN models by using a very large minibatch size $\mathcal{B}$. The large batch SGD is enabled by distributing the minibatch over $K$ devices, where a minibatch of $\frac{\mathcal{B}}{K}$ training examples is trained on each device. Therefore, the overall batch size $\mathcal{B}$ can be scaled up linearly with the number of devices $K$. However, training with a large batch size typically increases the generalization gap (i.e., a degraded test accuracy) \cite{keskar2016large, largebatch1, largebatch2, largebatch3,goyal2017accurate,wu2018group,hoffer2017train,smith2017don,peng2018megdet}. In other words, increasing batch size often leads to a significant loss in test accuracy. There has been a significant amount of effort in addressing the generalization gap. \citet{keskar2016large} provide a comprehensive analysis of large minibatch training for DNNs, which demonstrates that the lack of generalization is caused by the fact of converging to a sharp minimizer using large minibatch SGD. Moreover, converging to sharp minimizer using large minibatches is likely a common case in training complex DNNs due to their loss landscape. Recent works mainly focus on two directions to address the generalization gap -- (1) controlling the learning rate has been extensively studied, such as linear scaling learning rate \cite{krizhevsky2014one}, sharp decaying \cite{zagoruyko2016wide}, dynamic scaling \cite{zhao2019adaptive}; (2) leveraging gradient noise for large minibatch training \cite{zhou2018,welling2011bayesian,smith2017bayesian,anisotropic_noise, wen2019empirical}. While those works demonstrate promising performance boost in large minibatch training at different scenarios, we found limited literature in addressing the generalization gap using SGD regularization algorithms, which directly improves the gradient generalizability of each batch. More importantly, a general regularization algorithm is independent to the aforementioned techniques, which can be combined together to further improve the generalization. As the proposed distinctive regularization is orthogonal to the aforementioned techniques, we believe that the distinctive regularization can be used together to further minimize the test accuracy degradation caused by a large batch size.


\textbf{Regularization and Contrast Learning} Adding regularizer to the total loss can help to promote a specific structure on the solution. For example, $\ell_2$ regularization is used to control the weight norm and improve generalization \cite{Krogh1991}, $\ell_1$ and group $\ell_1$ regularization impose  (group) sparsity structure on the model parameters \cite{Tibshirani2011,Simon2013}, and nuclear norm regularization constraints the high dimensional parameters within a low dimensional subspace \cite{candes2012,Toh2010}. Other approaches regularize the algorithm update rules, e.g., the cubic regularization algorithm \cite{nesterov2006,zhou2018,wang19d}. All these conventional regularization approaches do not change the model structure. 
On the other hand, contrast learning \cite{hadsell2006dimensionality} has been introduced and demonstrated to create invariance of the input-output mapping, which improves generalization. For example, Momentum Contrast(MoCo) \cite{he2020momentum} introduces momentum encoder to train target encoder using contrast loss during pre-training phase, which aims to split the mapping space by clustering metrics of same class while pushing away different ones. Similar contrast loss has also been applied to helps enlarge the margin of different classifiers to improve the predication accuracy in few-shot learning \cite{wang2018large}. Inspired by the contrast loss \cite{hadsell2006dimensionality}, our DReg approach takes advantages of positive and negative keys to make contrast \cite{he2020momentum}, where DReg introduces contrast loss on the training parameters with replicated layers that largely share the rest model. Thus, DReg is expected to extract different but more generalizable features between the two input-output mappings.


\section{Approach}\label{sec:approach}

Convolution neural network (ConvNet) is a popular deep learning model that has been widely applied in many deep learning tasks.
Stochastic gradient descent (SGD) has demonstrated superior performance in training ConvNet at various scales.
While recent works attempt to close the generalization gap of training with a large batch size, existing approaches focus on adaptive scaling of gradients or learning rates and adding experimental noise. However, such approaches do not generalize the information during training, i.e., gradients. Thus, we propose a novel regularization approach, namely \textit{distinctive regularization} (DReg), which aims to directly generalize the propagated gradients. The main idea of DReg is introducing a new regularization in addition to the original classification loss, which improves the generalizability at each training step.



  \pdfoutput=1
  
  \begin{figure}[t]
  \centering
    \begin{subfigure}{0.42\textwidth}
      \centering
      \includegraphics[width=1.\linewidth]{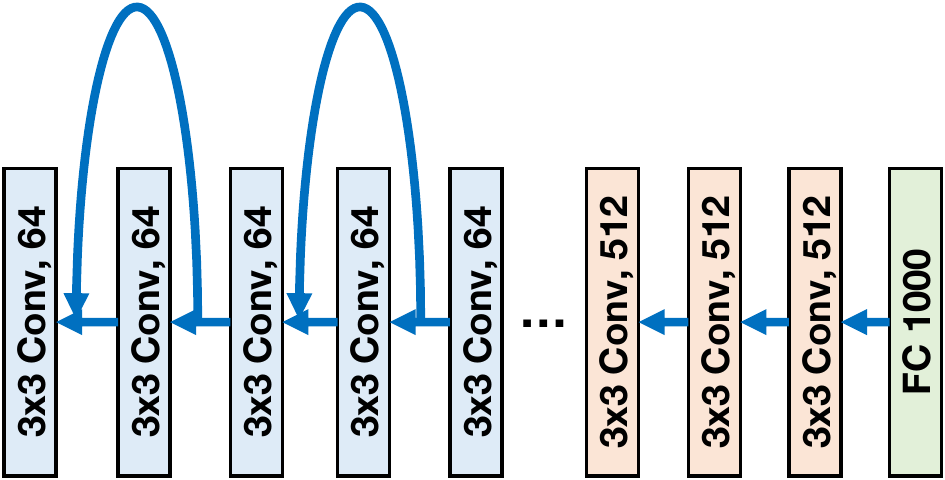}
      \caption{Backpropogation path of vanilla SGD where output gradient equals to $\partial \mathcal{L}(x,W)$.}
      \label{fig:overview_1}
    \end{subfigure}%
    \hspace{5mm}
    \begin{subfigure}{0.42\textwidth}
      \centering
      \includegraphics[width=1.\linewidth]{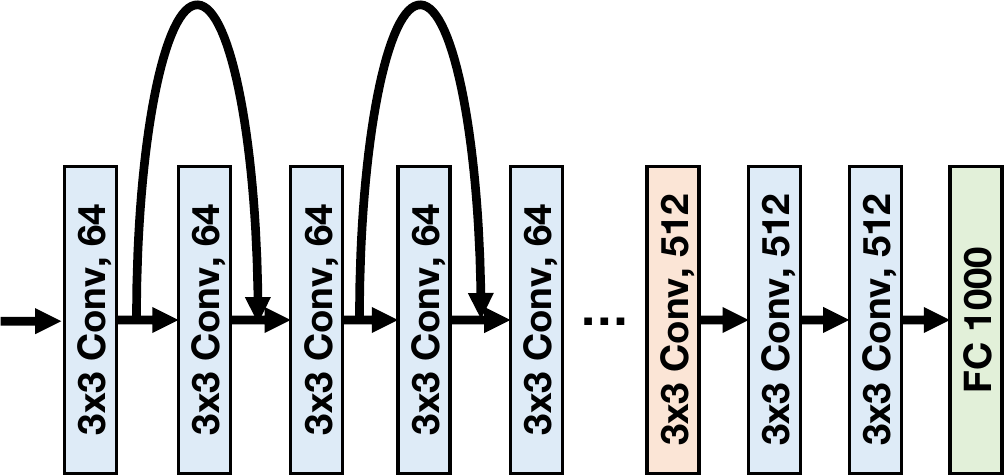}
      \caption{Final model (inference) with vanilla SGD.}
      \label{fig:overview_2}
    \end{subfigure}
    
    \begin{subfigure}{0.42\textwidth}
      \centering
      \includegraphics[width=1\linewidth]{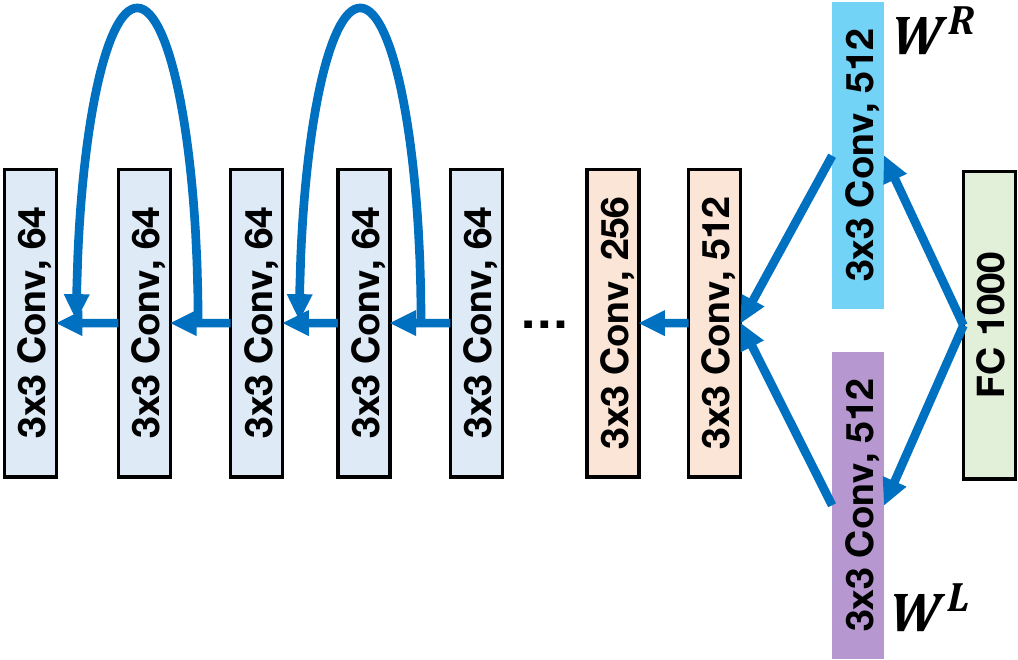}
      \caption{Backpropogation path of SGD with DReg where loss function is $\mathcal{L}(x,W, W^R) + \mathcal{L}(x,W, W^L) - \lambda \cdot \mathcal{L}_{\text{DReg}}(x,W^R, W^L)$.}
      \label{fig:overview_3}
    \end{subfigure}%
    \hspace{5mm}
    \begin{subfigure}{0.42\textwidth}
      \centering
      \includegraphics[width=1.\linewidth]{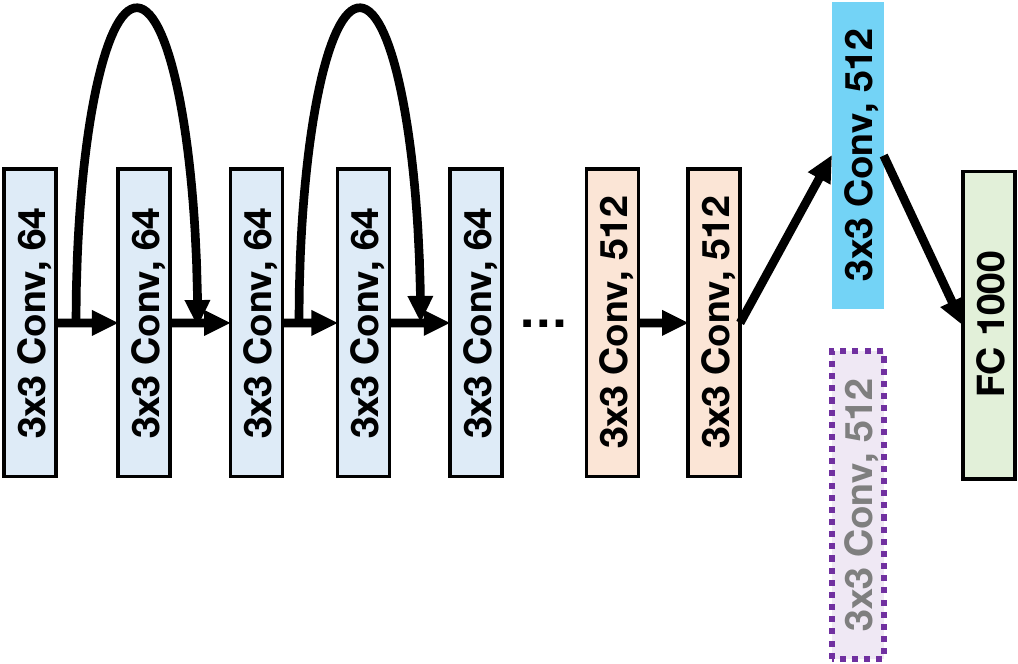}
      \caption{Final model (inference) with distinctive-regularized SGD.}
      \label{fig:overview_4}
    \end{subfigure}
    \caption{Illustration of the proposed distinctive regularization (DReg) approach using ResNet architecture -- \textbf{(a)} Layer-level computation graph of vanilla SGD backpropogation where $\mathcal{L}$ is a cross-entropy loss; \textbf{(b)} Final model (inference) with vanilla SGD.; \textbf{(c)} Layer-level computation graph of distinctive-regularized SGD backpropogation where loss function includes cross-entropy loss $\mathcal{L}$, and DReg loss $\mathcal{L}_{\text{DReg}}$ which aims to increases the distinctiveness of $W^R$ and $W^L$; \textbf{(d)} Final model (inference) with distinctive-regularized SGD is identical to vanilla SGD training by selecting the path with highest validation accuracy.}
    \label{fig:overview}
    \end{figure}    

\textbf{Distinctive Regularization (DReg)} We present our DReg method along with an illustrative example of ResNet architecture (Figure \ref{fig:overview}). Figures \ref{fig:overview_1} and \ref{fig:overview_2} illustrate the computation graph of the backpropagation and forward computation trained with the vanilla SGD, with a cross-entropy loss function $\mathcal{L}$.
As depicted in Figure \ref{fig:overview_3}, we introduce a new layer type called DReg layer, which consists of two Conv layers in parallel parameterized by $W^R$ and $W^L$.
Let $W$ be the weights of the layers, excluding the DReg layer. The proposed DReg method modifies the vanilla SGD in two aspects: \textbf{(1)} a novel dual backward paths (Figure \ref{fig:overview_3}) constructed at the selected layer(s) (DReg layer); \textbf{(2)} a novel loss function consists of cross-entropy loss $\mathcal{L}$ and DReg loss $\mathcal{L}_{\text{DReg}}$, where $\mathcal{L}_{\text{DReg}}$ is a \textit{squared Frobenius norm} of {\scriptsize ${||W^R - W^L||}_F^{2}$} used for guiding the distinctiveness. 
%
%

The intuition of DReg comes from the following observation: for large-scale and complex non-convex tasks, there exists a large number of unique non-convex functions that perform the same (i.e., achieve the same validation loss). {Therefore, we propose to find two distinctive paths that both minimize the cross-entropy loss, with the hope that one path can provide the momentum for the other path that is stuck in a local minima.} \textbf{By introducing the DReg layer and the DReg loss during backpropagation, the proposed DReg aims to search for a common $W$ such that the two forward paths with $W^R$ and $W^L$ can achieve low prediction loss while maximizing the difference between $W^R$ and $W^L$. 
In addition, introducing the DReg regularization can also improve the robustness/generalizability of $W$, since SGD with the DReg loss optimizes both paths with respect to the prediction accuracy, as well as optimizes $W$ to be robust for both $W^R$ and $W^L$ that are distinctive.} 
Finally, the forward path trained with DReg will be the same as the vanilla SGD by selecting the path with the highest validation accuracy (Figure \ref{fig:overview_4}).

\textbf{Loss Function} Let $W^*$ be the weights of the entire model, where $W^*=$ \{$W$, $W^R$, $W^L$\}. Let $f(\cdot; W, W^R)$ and $f(\cdot; W, W^L)$ be two independent prediction functions of given dataset $X$. Let the batch size be $M$, the loss function is defined as follows:
{
\vspace{-1mm}
\begin{equation}
    \mathcal{L} = \frac{1}{M} \sum_{x \in X} \overbrace{l(x, W, W^R)}^{\mathcal{L}_R} + \overbrace{l(x, W, W^L)}^{\mathcal{L}_L} - \overbrace{\lambda\cdot{||W^R - W^L||}_{F}^2}^{\mathcal{L}_{\text{DReg}}}
    \label{eq:loss}
\end{equation}
}
To train the ConvNet with DReg, the SGD optimizer is used to minimize the loss function $\mathcal{L}$. In order to minimize $\mathcal{L}$, the optimizer minimizes $\mathcal{L}_R$ and $\mathcal{L}_L$, and maximizes $\mathcal{L}_{\text{DReg}}$. 
We can see that minimizing $\mathcal{L}_R$ and $\mathcal{L}_L$ will optimize the prediction performance of both $f(\cdot; W, W^R)$ and $f(\cdot; W, W^L)$. Meanwhile, maximizing $\mathcal{L}_{\text{DReg}}$ will increase the distinctiveness of $W^R$ and $W^L$. While $\mathcal{L}_{\text{DReg}}$ is being maximized, the decreasing value of $\mathcal{L}_R$ and $\mathcal{L}_L$ will lead the SGD process to a more generalized $W$, which improves the performance of the overall model.
While Dreg loss can be constructed with more than one layer, we consider DReg loss with only a single layer in this work.

\textbf{DReg Configurations and Initialization} There are two main hyperparameters for constructing DReg -- \textbf{(1)} the scaling factor of the DReg loss $\lambda$; \textbf{(2)} the selected layer for constructing the DReg loss. We provide comprehensive evaluations of the impacts of these two parameters in Section \ref{sec:results} on various datasets using state-of-the-art ConvNet models. Regarding the initialization of $W^R$ and $W^L$, they are randomly initialized to be slightly distinctive, i.e., keeping the $\mathcal{L}_{\text{DReg}}$ small but greater than zero. This initialization method is developed empirically.

\begin{algorithm}[t]
\SetAlgoLined
\KwResult{$\boldsymbol{W}$=\{$\boldsymbol{W_0}$, $\boldsymbol{W_1}$, $\boldsymbol{W_2}$, ...\}}
 $\boldsymbol{\partial_{i}} \leftarrow$ $\partial\mathcal{L}_R + \partial\mathcal{L}_L - \partial\mathcal{L}_{\text{DReg}}$\; 
 \For{i $\leftarrow$ nodes of the computation graph in reversed topological order}{
  \eIf{$i = \boldsymbol{i_{\text{DReg}}}$}{
     $\boldsymbol{W_i^L} \leftarrow \boldsymbol{W_i^L} -\boldsymbol{\partial \mathcal{L_L} \slash \partial {W_i^L}}+\boldsymbol{\partial {||W_i^L-W_i^R||}_F^2} \slash \boldsymbol{\partial W_i^L}$\;
   $\boldsymbol{W_i^R} \leftarrow \boldsymbol{W_i^R} - \boldsymbol{\partial \mathcal{L_R} \slash \partial {W_i^R}} +\boldsymbol{\partial {||W_i^L-W_i^R||}_F^2} \slash \boldsymbol{\partial W_i^R}$\;
   $\boldsymbol{\partial_{i}}$ $\leftarrow $ \texttt{Backprop}($\boldsymbol{W_i^R}$, $\boldsymbol{\partial_{i+1}}$)$+$\texttt{Backprop}($\boldsymbol{W_i^L}$, $\boldsymbol{\partial_{i+1}}$)\;
   }{
   $\boldsymbol{W_i}$ $\leftarrow \boldsymbol{W_i} -$  \texttt{GradW}($\boldsymbol{\partial_{i}}$)\;
   $\boldsymbol{\partial_{i}}$ $\leftarrow$ \texttt{Backprop}($\boldsymbol{W_i}$, $\boldsymbol{\partial_{i+1}}$)
  }
 }  
 \caption{SGD with distinctive regularization (DReg)}
\end{algorithm}


\textbf{Gradient Updates} We present the complete gradient descent procedure of DReg in Algorithm 1. According to Equation \ref{eq:loss}, the loss function $\mathcal{L}$ consists of three parts: cross-entropy loss $\mathcal{L}_R$ of $f(\cdot; W, W^R)$, cross-entropy loss $\mathcal{L}_L$ of $f(\cdot; W, W^L)$, and DReg loss $\mathcal{L}_{\text{DReg}}$. Thus, for a given batch, the algorithm will initialize the output gradient as $\partial\mathcal{L}_R + \partial\mathcal{L}_L - \partial\mathcal{L}_{\text{DReg}}$ (line~1). Similarly to standard SGD, the gradients are propagated through the computation graph in a reversed topological order (line~2). Let $i_{\text{DReg}}$ be the index of the DReg layer, and $i$ is current layer with gradient propagated. \texttt{GradW} is the function that calculates the updates of the corresponding cross-entropy loss, e.g., \texttt{GradW} of $W^L$ path is the update from $\mathcal{L}_R$. While the gradient propagates to the DReg layer ($i=i_{\text{DReg}}$), our algorithm will update $W_i^{L}$ and $W_i^{R}$ with updates from both cross-entropy loss and DReg loss (lines~3-6). For the next node in the computation graph, both gradients from $W^R$ and $W^L$ will be propagated (line~6). The backpropagation procedure remains the same for the rest of nodes (lines~8-9). We analyze the two-step recursive gradient update of $W^R$ for the distinctive-regularized SGD to explain how DReg increases the generalizability of the gradient (Equations \ref{eq:gradient_update_explain_1}-\ref{eq:gradient_update_explain}). In Equation \ref{eq:gradient_update_explain}, two additional terms compared to vanilla SGD can be observed (updating a single path). As distinctive-regularized SGD proceeds, the first extra term increases the \textit{distinctiveness} between $W^R$ and $W^L$. Moreover, the second term introduces more diverse gradient updates, which will become more \textit{diverse} as the \textit{distinctiveness} stochastically increases -- as ${||W^R-W^L||}_{F}$ increases, the gradients of a given batch becomes more diverse and generalizable since {\scriptsize $\triangledown{W^R_{t-1}}\mathcal{L}_R$} and {\scriptsize$\triangledown{W^L_{t-1}}\mathcal{L}_L$} become more diverse. More importantly, the gradient updates of $W^R$ involves graidents of its distintive $W^L$ path.   

{
\vspace{-2mm}
\footnotesize  
    \begin{align}
     &W_{t+1}^R = W^R_{t} - \eta \triangledown_{W^R_{t}}\mbox{\scriptsize $\mathcal{L}_{R}$} - \eta \triangledown_{W^R_{t}}\mbox{\scriptsize $\mathcal{L}_{L}$} + {\lambda\eta \triangledown_{W^R_{t}}{||W_t^R - W_t^L||}_F^2}
     \label{eq:gradient_update_explain_1}\\
     &=W^R_{t} - \eta \triangledown_{W^R_{t}}\mbox{\scriptsize $\mathcal{L}_{R}$} - 0 + \mathbf{2\lambda\eta(W_{t}^R - W_{t}^L)}\\
     &\mathbf{W_{t}^R - W_{t}^L}=(W_{t-1}^R - \eta \triangledown_{W^R_{t-1}}\mbox{\scriptsize $\mathcal{L}_{R}$} +  \lambda \eta \triangledown_{W^R_{t-1}}\mbox{\scriptsize $\mathcal{L}_{DReg}$}) - (W_{t-1}^L - \eta \triangledown_{W^L_{t-1}}\mbox{\scriptsize $\mathcal{L}_{L}$}+\lambda\eta\triangledown_{W^L_{t-1}}\mbox{\scriptsize $\mathcal{L}_{DReg}$})\\
     &=(W_{t-1}^R - W_{t-1}^L) - \underbrace{ \lambda \eta(\triangledown_{W_{t-1}^R} \mbox{\scriptsize $\mbox{\scriptsize $\mathcal{L}_{DReg}$}$} - \triangledown_{W_{t-1}^L}\mbox{\scriptsize $\mathcal{L}_{DReg}$})}_{\text{\small distinctiveness}} - \underbrace{\eta (\triangledown_{W^R_{t-1}}\mbox{\scriptsize $\mathcal{L}_{R}$}-\triangledown_{W^L_{t-1}}\mbox{\scriptsize $\mathcal{L}_{L}$})}_{\text{\small gradient~diversity}}
     \label{eq:gradient_update_explain}
    \end{align}
}

\textbf{Computational Training Overhead} The computational overhead of DReg is marginal compared to vanilla SGD, in particular for ConvNets in distributed training system. The overhead comes from two sources -- \textbf{(1)} \textit{Forward path}. As shown in Figure \ref{fig:overview_3}, DReg loss involves only a single Conv layer that closes to the output layer, and the rest of the forward path remains the same. Hence, the extra computations include forward computation through DReg layer and the output layer. \textbf{(2)} \textit{Backward path}. Similarly, as shown in Algorithm 1, the extra computations come from DReg layer updates for propagating $\mathcal{L}_{\text{DReg}}$, and the rest of the backward computations remain the same as vanilla SGD. 
\section{Experimental Results}\label{sec:results}

In this section, we empirically evaluate DReg on the CIFAR10 and CIFAR100 dataset~\cite{krizhevsky2014cifar} with a variety of ConvNets including VGG16 \cite{simonyan2014vgg}, MobileNetV2 \cite{sandler2018mobilenetv2}, and ResNet18 \cite{he2016deep}.
The main findings are as follows: -- \textbf{(a)} Section \ref{sec:convergence} presents the analysis of DReg in accelerating minibatch SGD, where we observe significantly convergence boost for large batch size (e.g., $2^{10}\cdot30$) across all experiments; \textbf{(b)} Section \ref{sec:dreg_position} provides intuitions for constructing DReg loss in different training scenarios; \textbf{(c)} Section \ref{sec:momentum} shows that DReg can accelerate SGD both with and without momentum. 

\textbf{General setup} 
We first train the original VGG16, MobileNetV2, and ResNet18 models using momentum SGD on CIFAR10 and CIFAR100 as the baseline.
We then empirically evaluate the large-batch training performance of distinctive-regularized momentum SGD for VGG16, MobileNetV2, and ResNet18 on CIFAR-10 and CIFAR100, with batch size varying from $2^{10}\cdot2$ to $2^{10}\cdot30$. By varying the DReg parameters, our experimental results on MobileNetV2 and ResNet18 provide intuitions for constructing DReg effectively. 

\textbf{DReg setup} {We construct DReg loss with $\lambda=0.1$ with the last 3$\times$3 Conv layer (DW-Conv) in all the models as DReg layer. 
In Section \ref{sec:dreg_position}, we analyze different hyperparamters in DReg, i.e., the topological position of DReg layer and the choices of $\lambda$.} \textit{Optimizer} -- We use the momentum SGD optimizer with default momentum 0.9 (Sections \ref{sec:convergence} and \ref{sec:dreg_position}). In Section \ref{sec:momentum}, we further explore the impact of DReg on different momentum setups. 

\subsection{Superior Convergence Boost for Large Minibatch SGD}\label{sec:convergence}

In this section, we demonstrate the effectiveness of DReg for accelerating minibatch SGD. Note that we do not aim to achieve the best validation accuracy for the given models and datasets. Instead, our focus is to demonstrate that distinctive-regularized momentum SGD can outperform momentum SGD in a variety of training setups.
    
\begin{figure}
    \begin{subfigure}{0.32\textwidth}
      \centering
      \includegraphics[width=1.\linewidth]{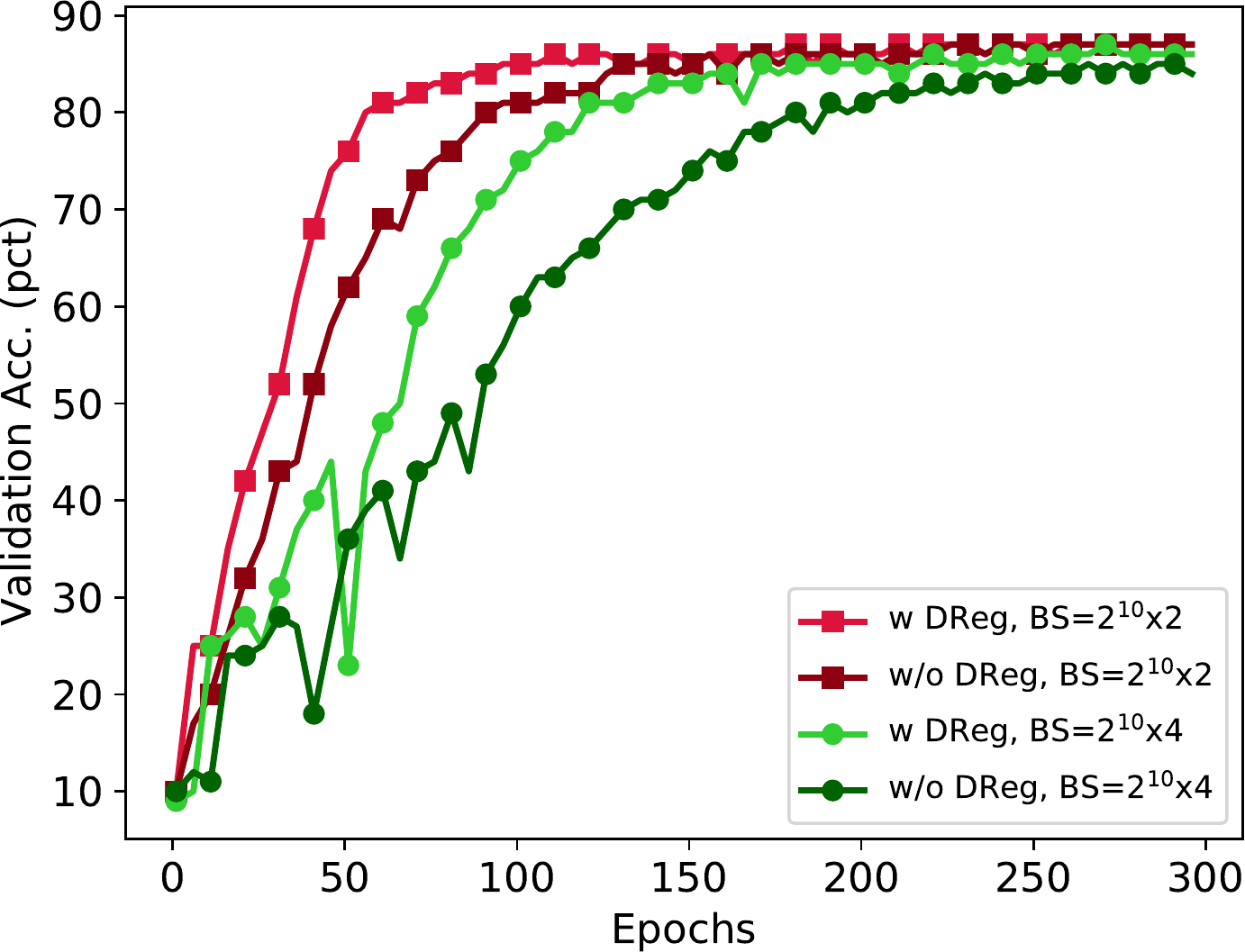}
      \caption{BS$\in \{2^{10}\cdot2$, $2^{10}\cdot4\}$}
      \label{fig:vgg16_cifar10_acc_1}
    \end{subfigure}%
    \begin{subfigure}{0.32\textwidth}
      \centering
      \includegraphics[width=1.\linewidth]{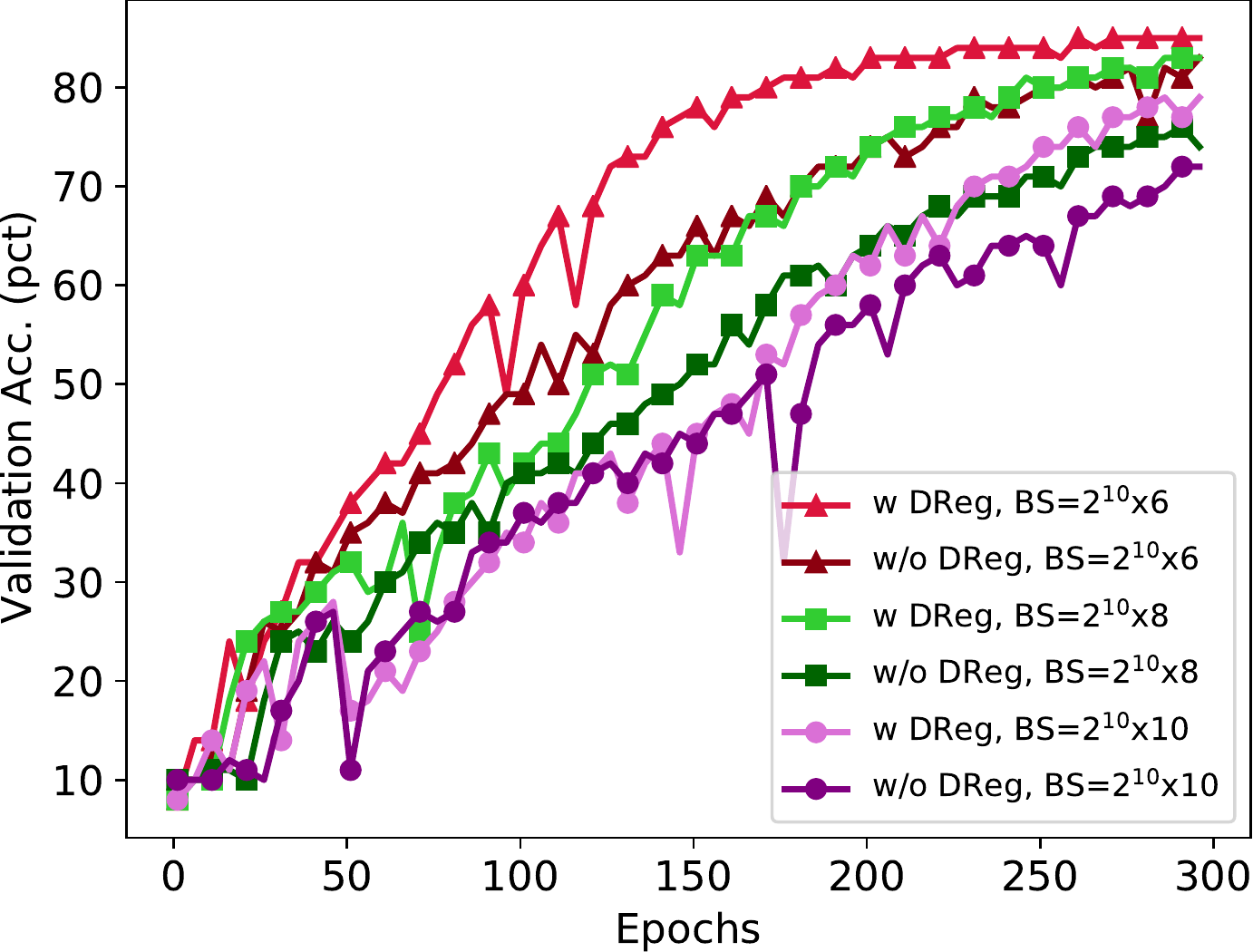}
      \caption{BS$\in \{2^{10}\cdot6$, $2^{10}\cdot8$,  $2^{10}\cdot10\}$}
      \label{fig:vgg16_cifar10_acc_2}
    \end{subfigure}
    \begin{subfigure}{0.32\textwidth}
      \centering
      \includegraphics[width=1.\linewidth]{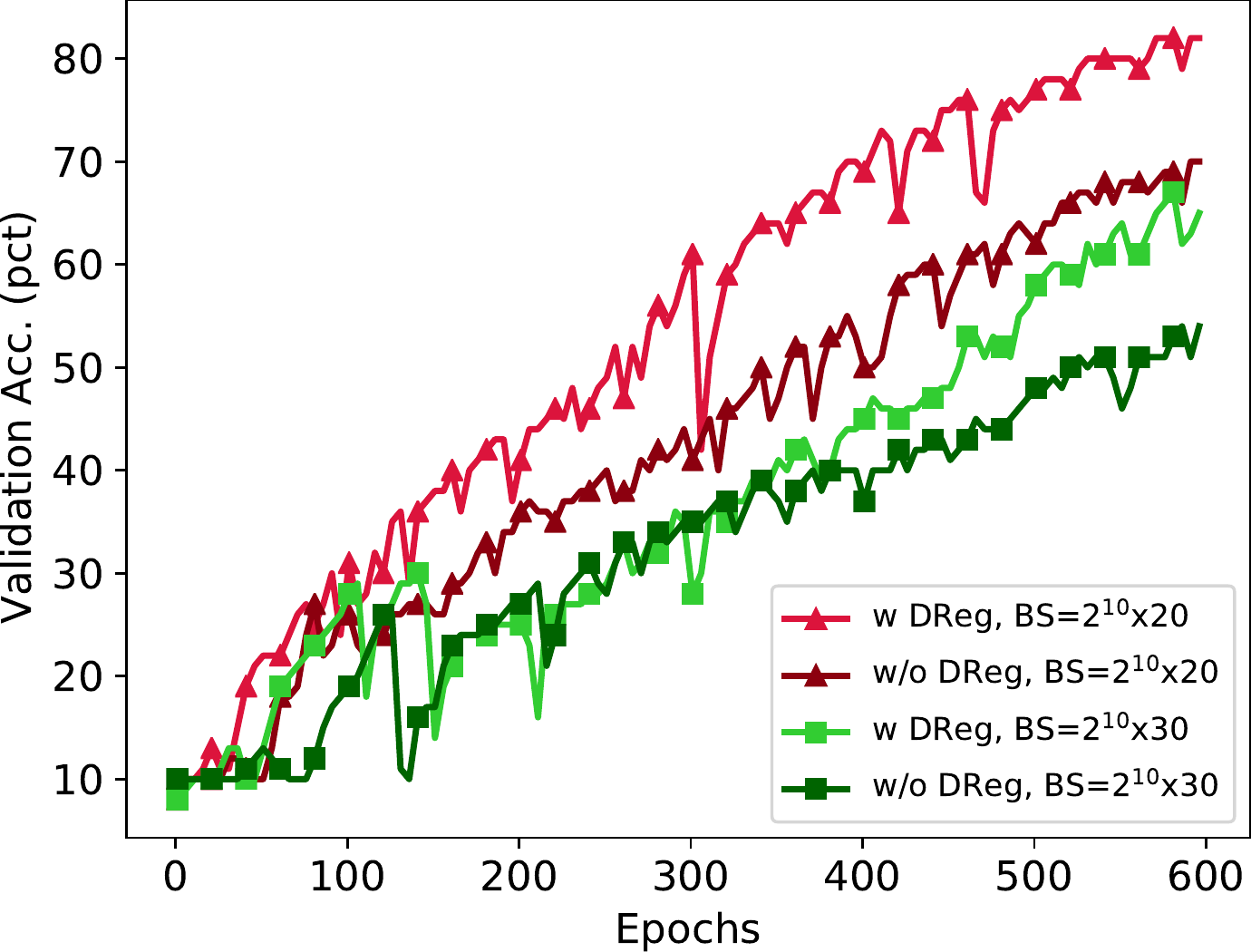}
      \caption{BS$\in \{2^{10}\cdot20$, $2^{10}\cdot30\}$}
      \label{fig:vgg16_cifar10_acc_3}
    \end{subfigure}
    \caption{Comparison results using VGG16 on CIFAR10 dataset with and without distinctive regularization (DReg) over a variety of batch sizes (BS) using momentum SGD algorithm, which demonstrate significant training acceleration boosted from DReg --- (a) BS$\in \{2^{10}\cdot2$, $2^{10}\cdot4\}$; (b) BS$\in \{2^{10}\cdot6$, $2^{10}\cdot8$,$2^{10}\cdot10\}$; (c) BS$\in \{2^{10}\cdot20$, $2^{10}\cdot30\}$.}
     \vspace{-2mm}
    \label{fig:vgg16_cifar10_acc}
    \end{figure}

\textbf{CIFAR10 Dataset} In Figure \ref{fig:vgg16_cifar10_acc}, we evaluate DReg on VGG16 with CIFAR10 dataset, using momentum SGD optimization algorithm with {$\lambda$=0.1}.
Specifically, we show the performance of DReg using various batch sizes (BS), including BS$\in 2^{10}\cdot$ \{$2,4$\} (Figure \ref{fig:vgg16_cifar10_acc_1}), BS$\in 2^{10}\cdot$\{$6,8,10$\} (Figure \ref{fig:vgg16_cifar10_acc_2}), and BS$\in 2^{10}\cdot$\{$20,30$\} (Figure \ref{fig:vgg16_cifar10_acc_3}). \textbf{First, compared to momentum SGD, we demonstrate that distinctive-regularized momentum SGD can significantly accelerate the training process for all batch sizes ranging from $2^{10}\cdot2$ to $2^{10}\cdot30$}. Compared to momentum SGD, distinctive-regularized momentum SGD can converge \textbf{2-3$\times$ faster} in terms of reaching the best validation accuracy.
For example, in Figure \ref{fig:vgg16_cifar10_acc_1}, while BS=$2^{10}\cdot4$, DReg takes 110 epochs to reach the highest validation accuracy, which can only be achieved after 300 epochs without DReg. \textbf{Second, as the batch size increases, the speedup of convergence using DReg increases}. In other words, DReg provides more significant acceleration in training with larger batch sizes. 
For instance, we can see that the validation accuracy gap between with and without DReg at each epoch is less than 10\% in Figure \ref{fig:vgg16_cifar10_acc_1}, where Figures \ref{fig:vgg16_cifar10_acc_2} and \ref{fig:vgg16_cifar10_acc_3} have up to 20\% validation accuracy difference.

   \pdfoutput=1
   
    \pdfoutput=1
    
    \begin{figure}
    \begin{subfigure}{0.32\textwidth}
      \centering
      \includegraphics[width=1.\linewidth]{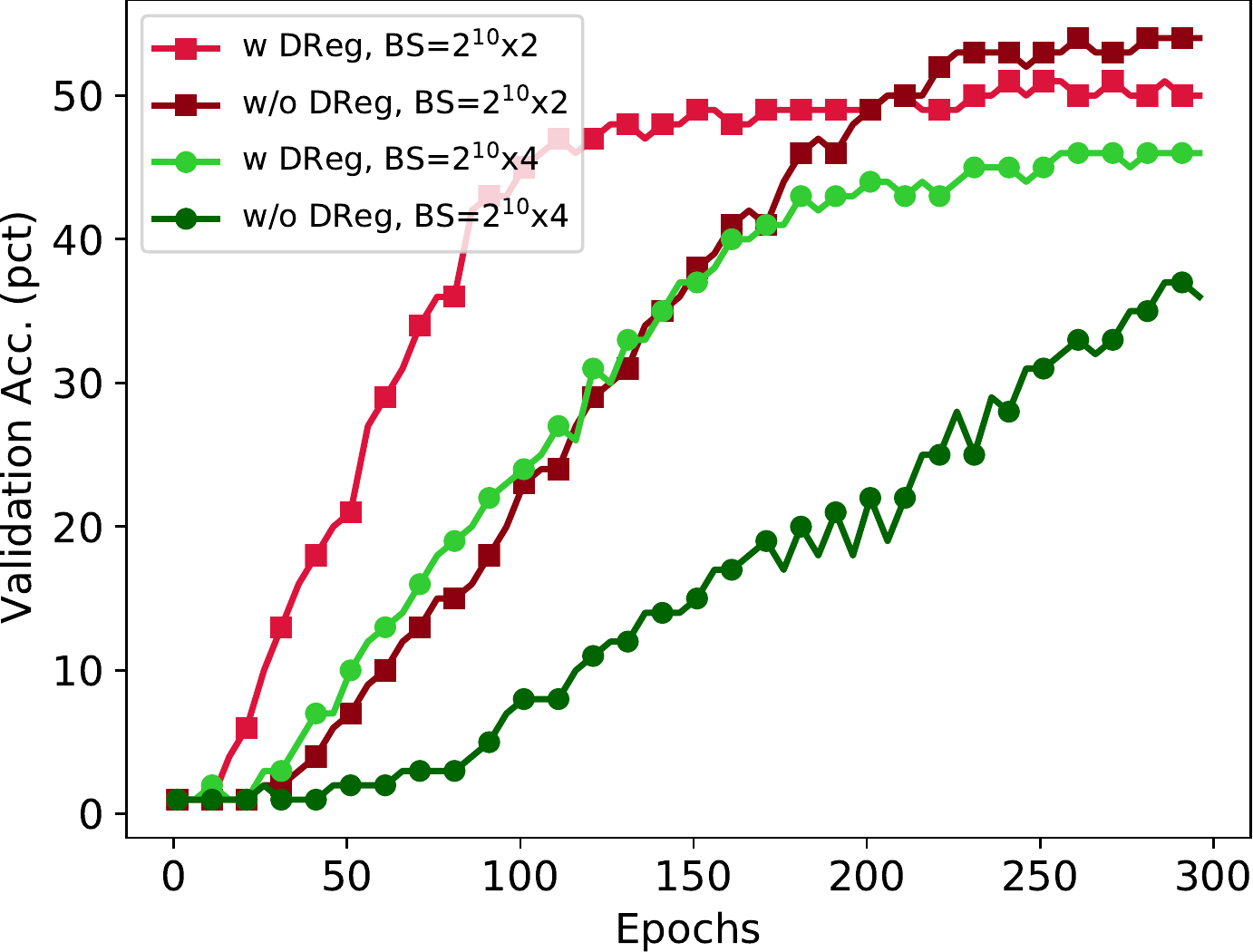}
      \caption{BS$\in \{2^{10}\times2$, $2^{10}\times4$\}}
      \label{fig:vgg16_cifar100_acc_1}
    \end{subfigure}%
    \begin{subfigure}{0.32\textwidth}
      \centering
      \includegraphics[width=1.\linewidth]{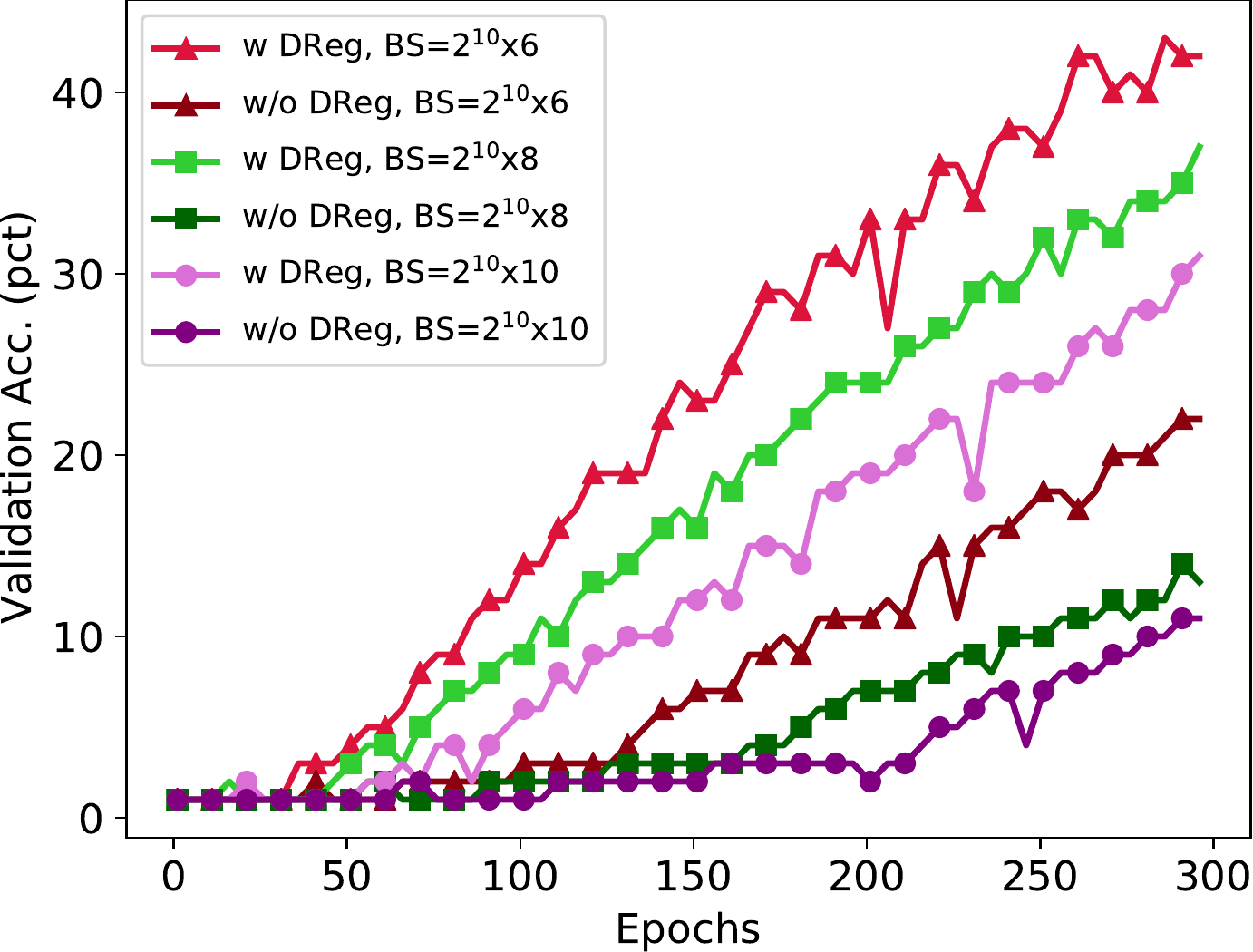}
      \caption{BS$\in \{2^{10}\times6$, $2^{10}\times8$,  $2^{10}\times10\}$}
      \label{fig:vgg16_cifar100_acc_2}
    \end{subfigure}
    \begin{subfigure}{0.32\textwidth}
      \centering
      \includegraphics[width=1.\linewidth]{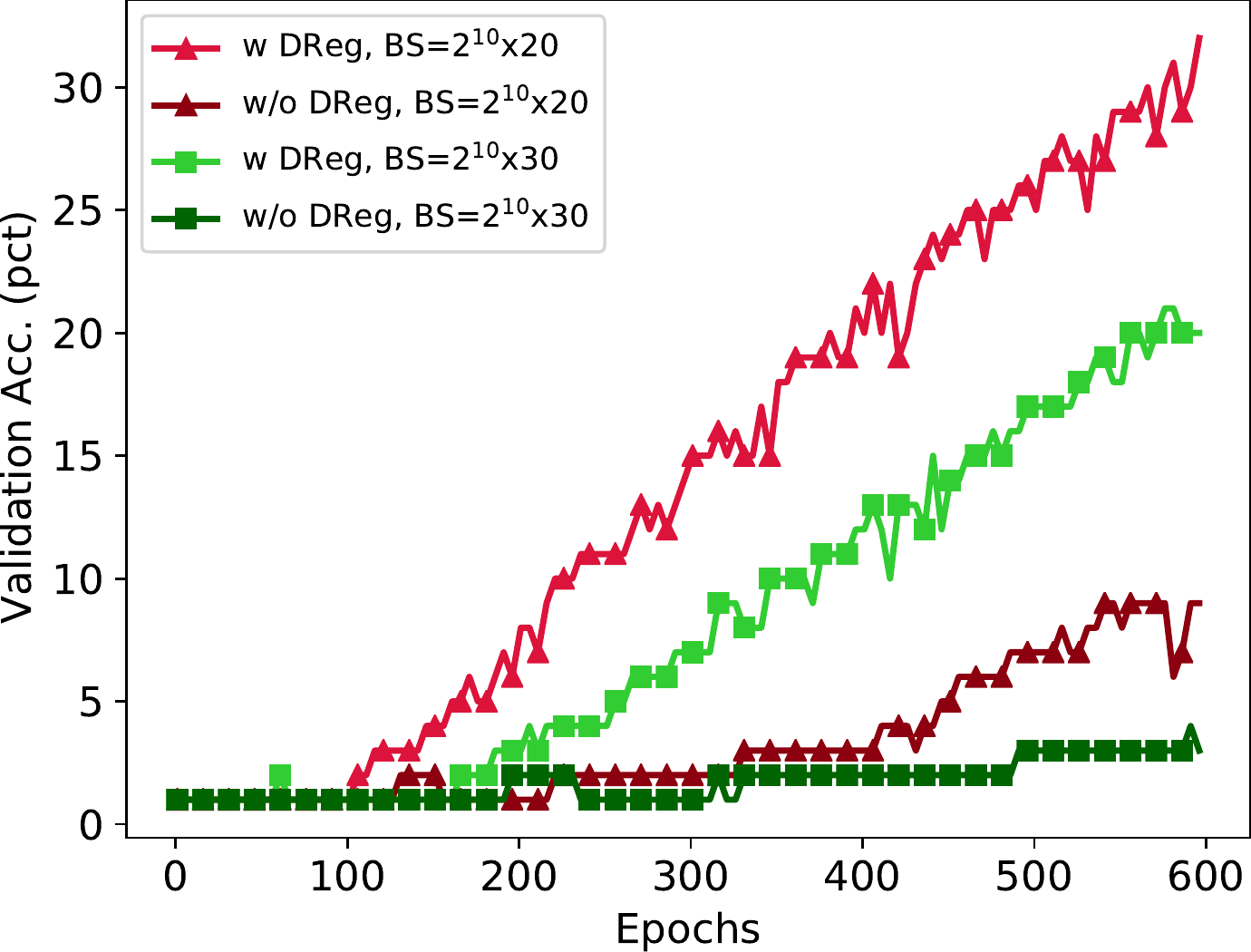}
      \caption{BS$\in \{2^{10}\times20$, $2^{10}\times30$\}}
      \label{fig:vgg16_cifar100_acc_3}
    \end{subfigure}
    \caption{Comparison results using VGG16 on CIFAR100 dataset w and w/o distinctive regularization (DReg) over a variety of batch sizes (BS) using momentum SGD algorithm, where DReg can further boost the training process compared to CIFAR10 dataset --- (a) BS$\in \{2^{10}\times2$, $2^{10}\times4\}$; (b) BS$\in \{2^{10}\times6$, $2^{10}\times8$,$2^{10}\times10\}$; (c) BS$\in \{2^{10}\times20$, $2^{10}\times30\}$.}
    \label{fig:vgg16_cifar100_acc}
    \vspace{-2mm}
    \end{figure}  

\textbf{CIFAR100 Dataset} To further demonstrate the effectiveness of DReg in accelerating training with large minibatch, we validate DReg on CIFAR100 dataset. 
First, using the same VGG16 model and training hyperparameters, we evaluate the performance of DReg on VGG16 with CIFAR100 dataset, as depicted in Figure \ref{fig:vgg16_cifar100_acc}. \textbf{Similarly, distinctive-regularized momentum SGD converges 2-3$\times$ faster than momentum SGD, and the convergence speedup increases as the batch size increases.} \textbf{Moreover, given the same model (VGG16), we observe that DReg can further boost the training process, particularly at the early stage of the training process.} For example, as shown in Figures \ref{fig:vgg16_cifar10_acc_2} and \ref{fig:vgg16_cifar10_acc_3}, the average validation accuracy gap over epochs with CIFAR10 dataset between with and without DReg is about $\sim$10\%. While for CIFAR100, the average validation accuracy gap is increased to $\sim$20 - 25\%.

\begin{figure}
    \centering
    \begin{subfigure}{0.32\textwidth}
      \centering
      \includegraphics[width=1.\linewidth]{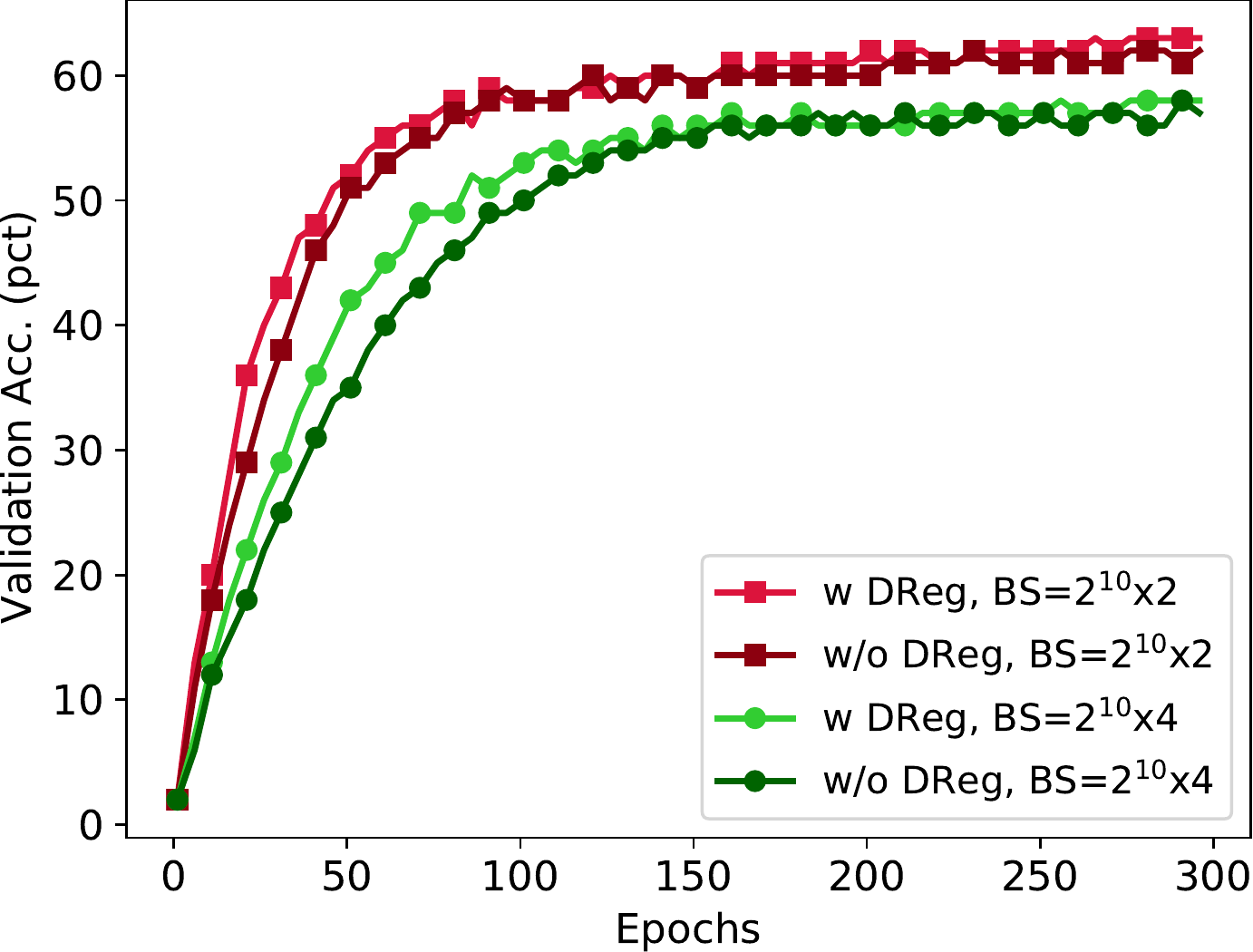}
      \caption{BS$\in \{2^{10}\cdot2$, $2^{10}\cdot4\}$}
      \label{fig:mobilenet_cifar100_acc_1}
    \end{subfigure}%
    \begin{subfigure}{0.32\textwidth}
      \centering
      \includegraphics[width=1.\linewidth]{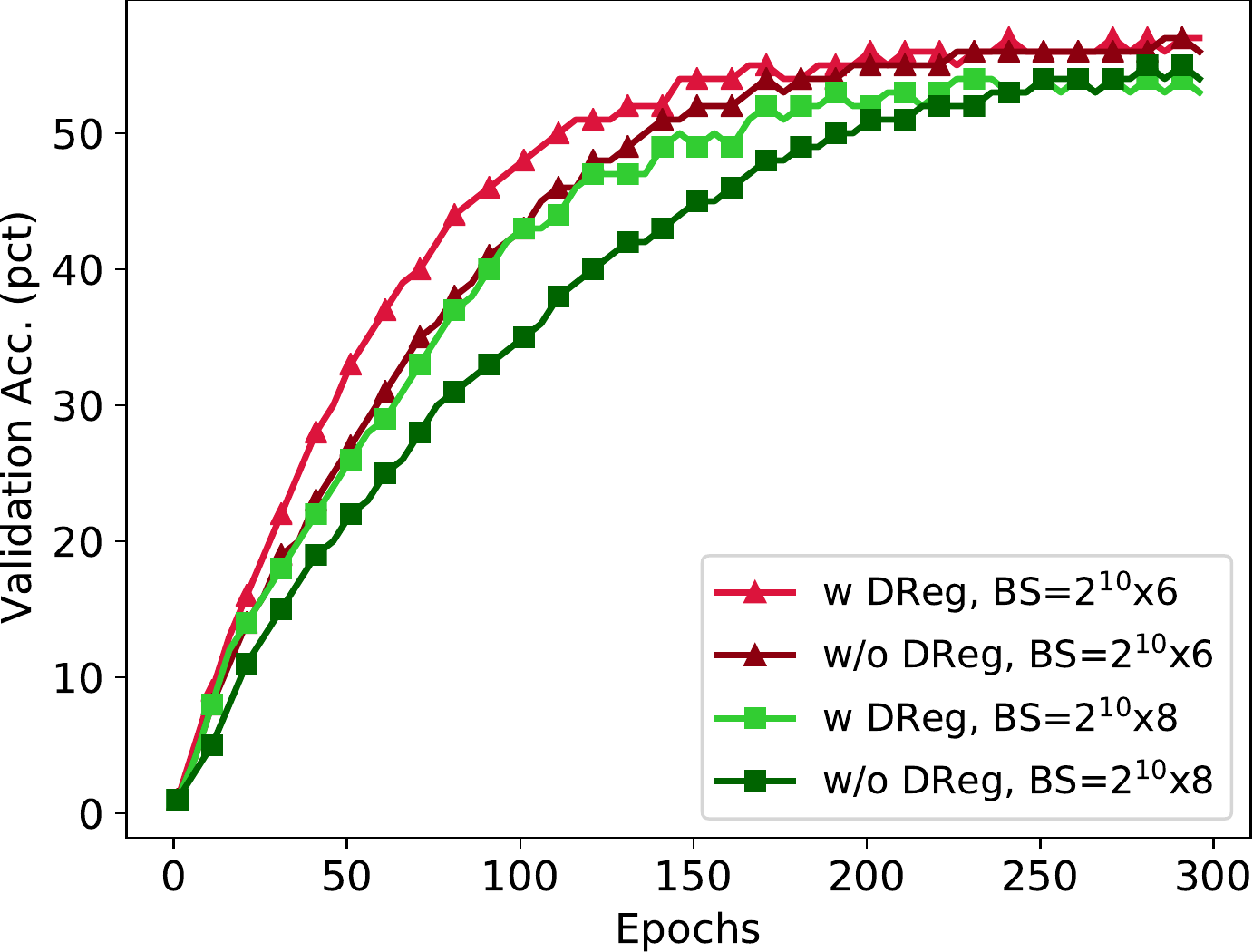}
      \caption{BS$\in \{2^{10}\cdot6$, $2^{10}\cdot8\}$}
      \label{fig:mobilenet_cifar100_acc_2}
    \end{subfigure}
    \begin{subfigure}{0.32\textwidth}
      \centering
      \includegraphics[width=1.\linewidth]{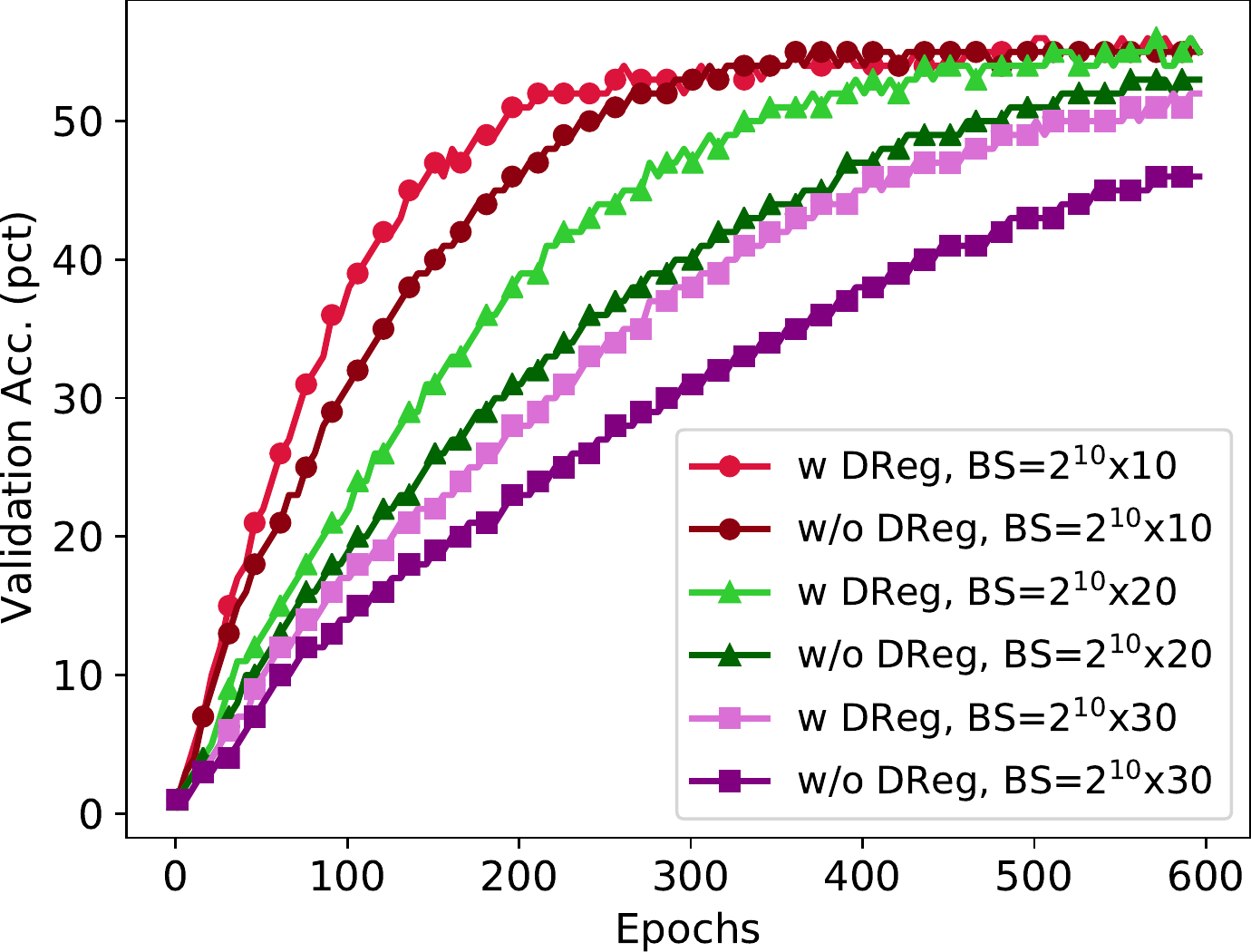}
      \caption{BS$\in \{2^{10}\cdot10$, $2^{10}\cdot20$, $2^{10}\cdot30\}$}
      \label{fig:mobilenet_cifar100_acc_3}
    \end{subfigure}
    \caption{Comparison results using MobileNetV2 on CIFAR100 dataset with and without distinctive regularization (DReg) over a variety of batch sizes (BS), using momentum SGD algorithm --- (a) BS$\in \{2^{10}\cdot2$, $2^{10}\cdot4\}$; (b) BS$\in \{2^{10}\cdot6$, $2^{10}\cdot8\}$; (c) BS$\in \{2^{10}\cdot10$, $2^{10}\cdot20$, $2^{10}\cdot30\}$.}
    \label{fig:mobilenet_cifar100_acc}
    \end{figure}
\begin{figure}
    \centering
    \begin{subfigure}{0.32\textwidth}
      \centering
      \includegraphics[width=1.\linewidth]{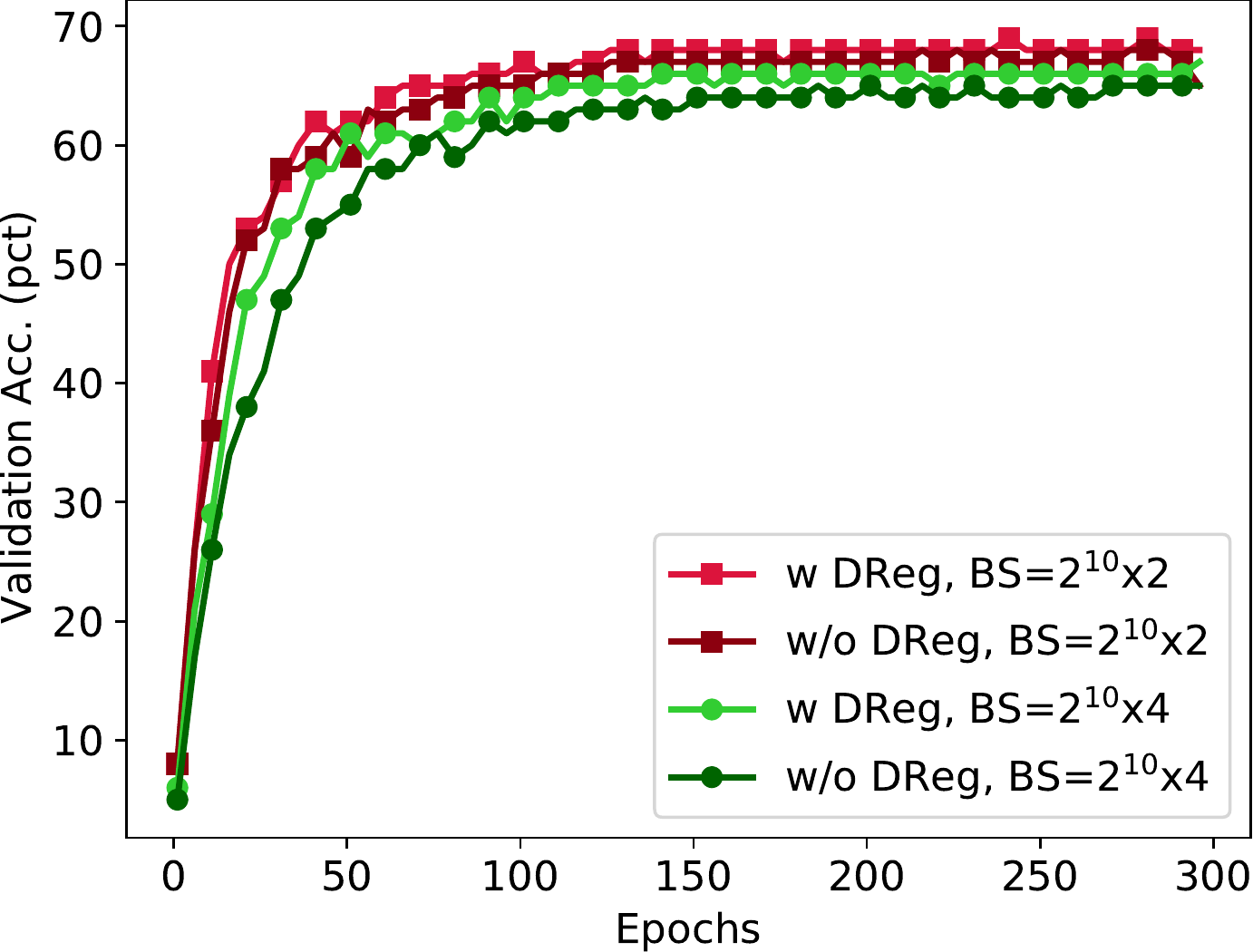}
      \caption{BS$\in \{2^{10}\cdot2$, $2^{10}\cdot4\}$}
      \label{fig:resnet18_cifar100_acc_1}
    \end{subfigure}%
    \begin{subfigure}{0.32\textwidth}
      \centering
      \includegraphics[width=1.\linewidth]{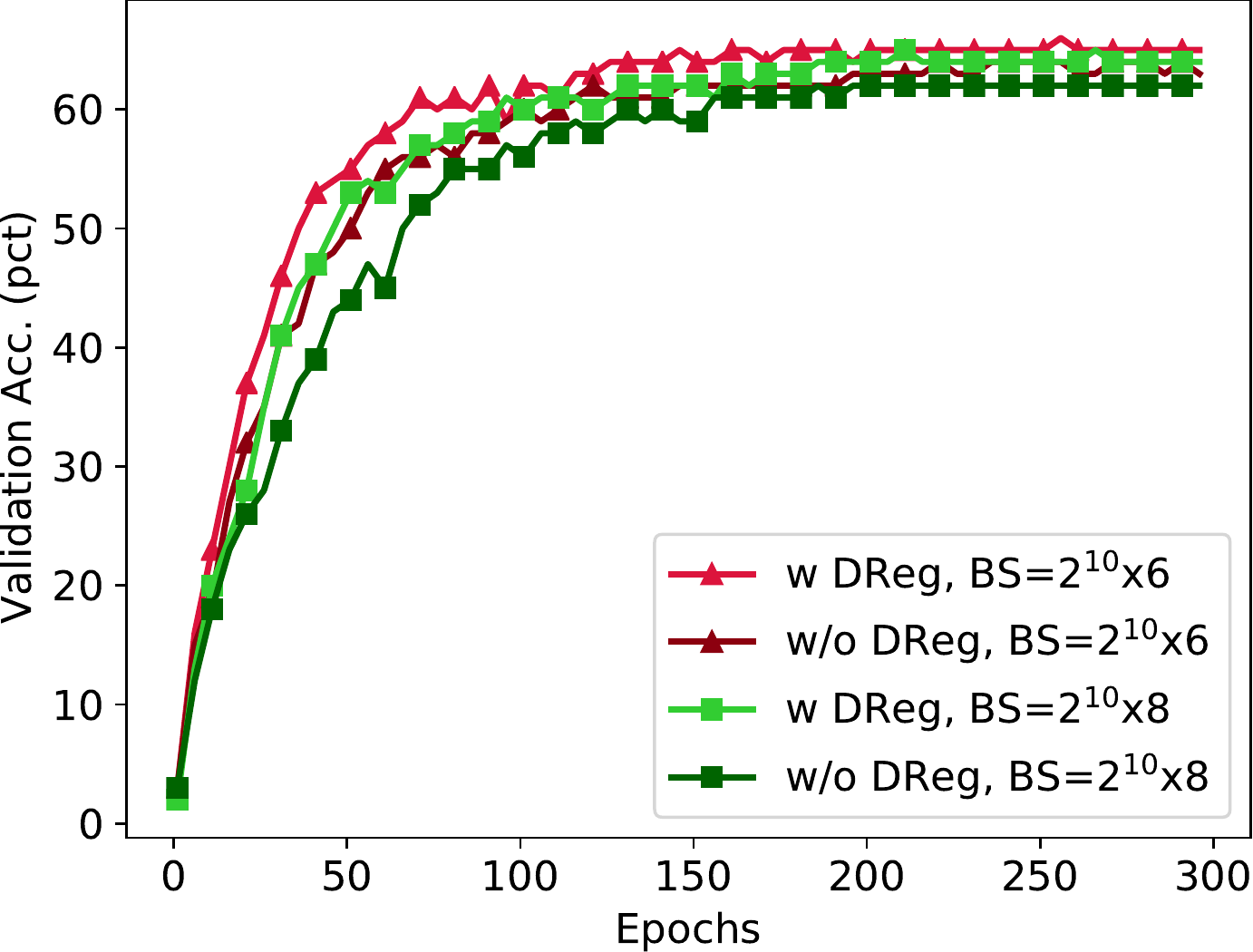}
      \caption{BS$\in \{2^{10}\cdot6$, $2^{10}\cdot8\}$}
      \label{fig:resnet18_cifar100_acc_2}
    \end{subfigure}
    \begin{subfigure}{0.32\textwidth}
      \centering
      \includegraphics[width=1.\linewidth]{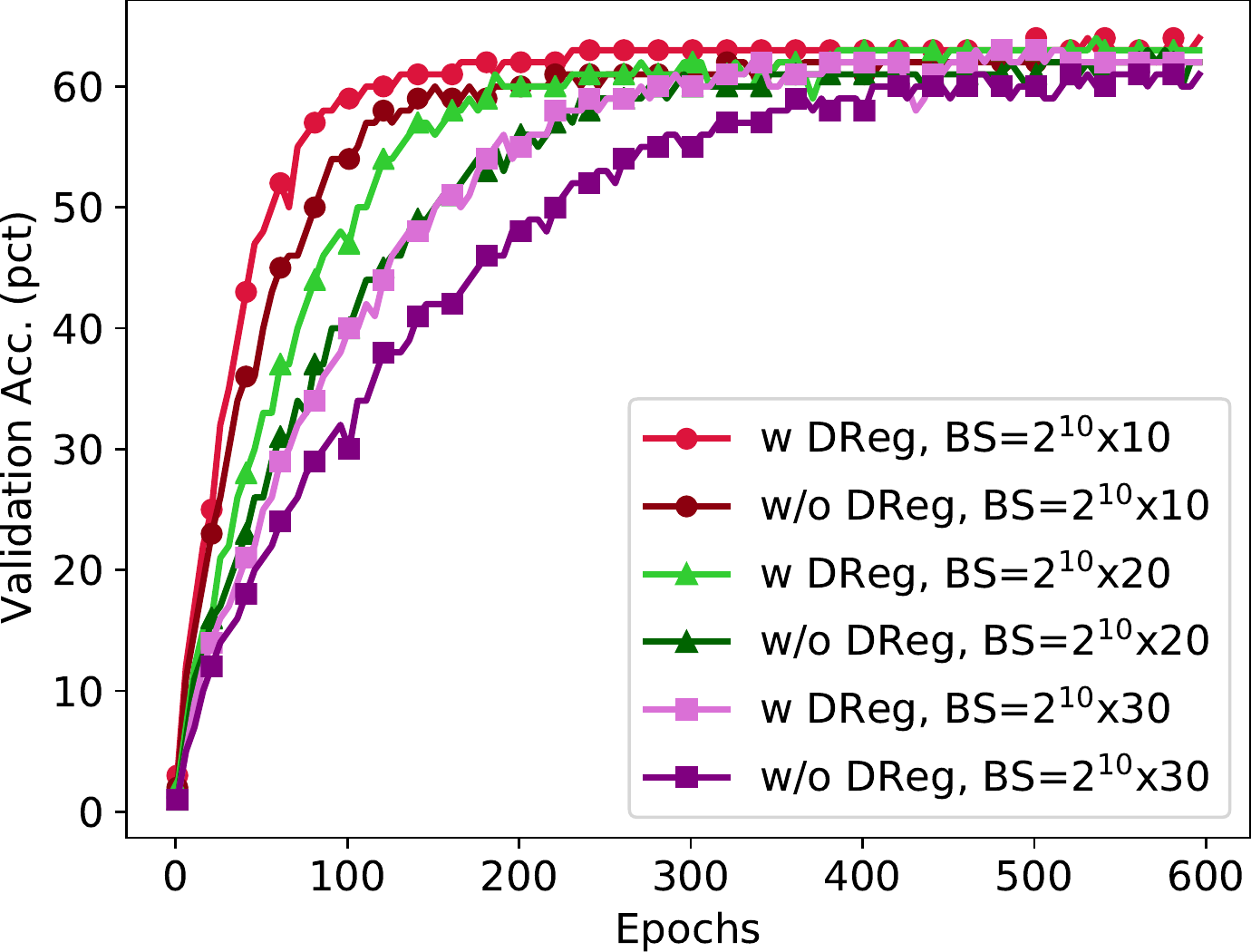}
      \caption{BS$\in \{2^{10}\cdot10$, $2^{10}\cdot20$, $2^{10}\cdot30\}$}
      \label{fig:resnet18_cifar100_acc_3}
    \end{subfigure}
    \caption{Comparison results using ResNet18 on CIFAR100 dataset with and without distinctive regularization (DReg) over a variety of batch sizes (BS), using momentum SGD algorithm --- (a) BS$\in \{2^{10}\cdot2$, $2^{10}\cdot4\}$; (b) BS$\in \{2^{10}\cdot6$, $2^{10}\cdot8\}$; (c) BS$\in \{2^{10}\cdot10$, $2^{10}\cdot20$, $2^{10}\cdot30\}$.}
    \label{fig:resnet18_cifar100_acc}
    \end{figure}

Moreover, we evaluate DReg on the state-of-the-art models such as MobileNetV2 and ResNet18, both of which contain a series of residual blocks. The results of MobileNetV2 and ResNet18 are presented in Figure \ref{fig:mobilenet_cifar100_acc} and \ref{fig:resnet18_cifar100_acc}, respectively. According to these results, for the more complex and robust models, DReg still offers a significant convergence speedup for batch sizes that are greater than $2^{10}\cdot6$, while the speedup is marginal for small batch sizes. Besides, we observe that the convergence speedup is increased with the batch size.
In conclusion, DReg outperforms the baseline in accelerating training convergence given a more complex dataset or a simper and less robust model or larger batch sizes.

\pdfoutput=1

   \pdfoutput=1
   
\subsection{Optimizing DReg Loss Construction -- Topological Sensitivity and the Choice of $\lambda$} \label{sec:dreg_position}

Based on the aforementioned DReg configuration (Section \ref{sec:approach}), we focus on analyzing the two hyperparameters involved in constructing DReg loss -- \textbf{(1)} The topological level where the DReg loss is introduced, i.e., the position of DReg layer in the neural network. \textbf{(2)} The hyperparameter $\lambda$ is used to scale the DReg loss versus the classification loss (cross-entropy loss). 

\begin{figure}[h!]
    \centering
    \begin{subfigure}{.36\textwidth}
      \centering
      \includegraphics[width=1\linewidth]{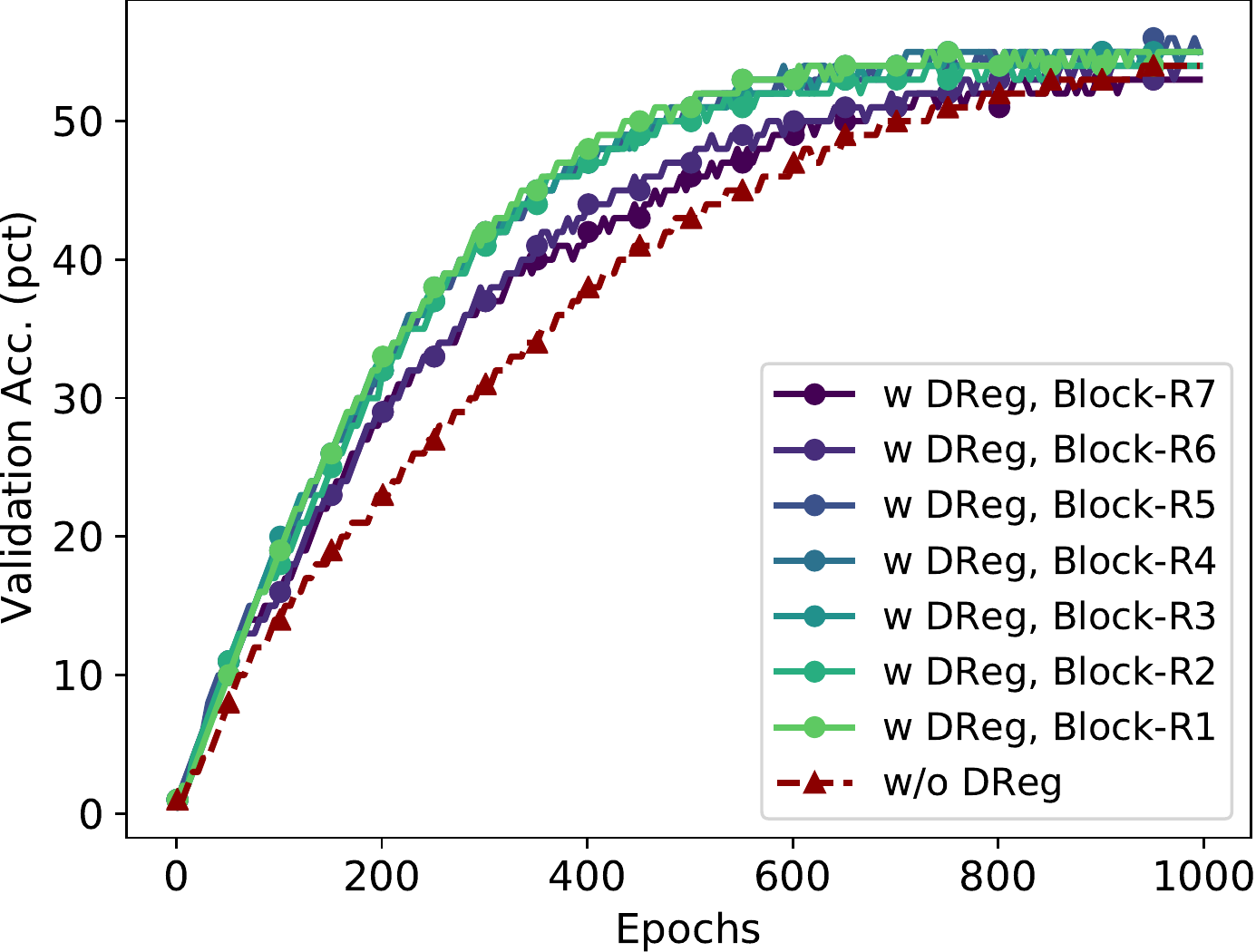}
      \caption{MobileNetV2}
      \label{fig:layer_selection_1}
    \end{subfigure}%
    \hspace{3mm}
    \begin{subfigure}{.36\textwidth}
      \centering
      \includegraphics[width=1\linewidth]{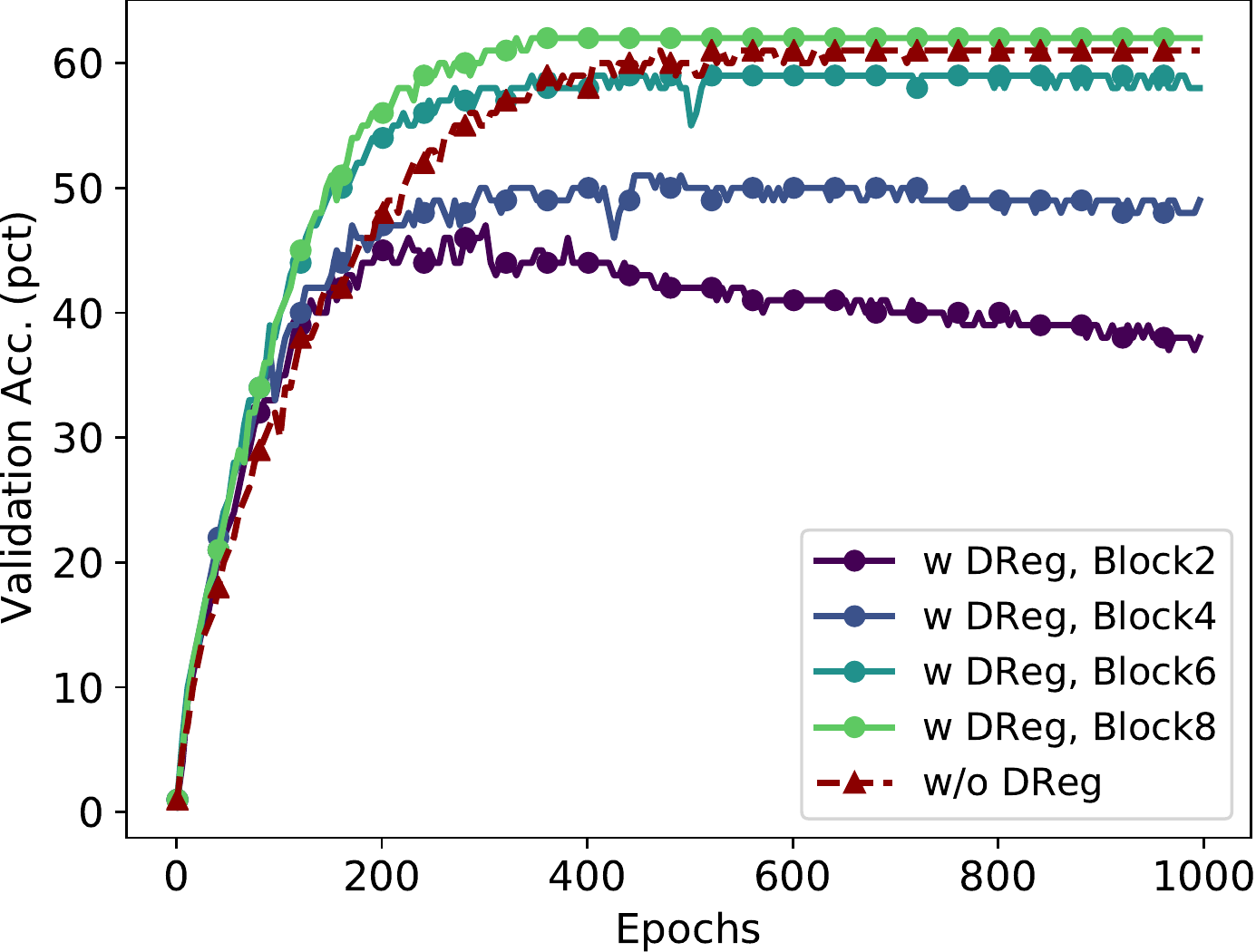}
      \caption{ResNet18}
      \label{fig:layer_selection_2}
    \end{subfigure}
    \caption{Analysis of building DReg at different topological levels using MobileNetV2 and ResNet18 architecture on CIFAR100 dataset with $\lambda =0.1$ and BS$=2^{10}\cdot30$. The selected depthwise (DW) convolutional layer used for constructing DReg loss is labeled in reverse topological order (from output to input), e.g., Block-R1 is the last DW-Conv of the model --- \textbf{(a)} Seven topological levels evaluated on MobileNetV2; \textbf{(b)} Four topological levels evaluated on ResNet18.}
    \label{fig:layer_selection}
    \end{figure}

\textbf{Topological Sensitivity Analysis} For the best of interests to image classification tasks, we evaluate the topological sensitivity using MobileNetV2 and ResNet18, which consist of residual blocks. The results are shown in Figure \ref{fig:layer_selection}, where we consider seven and four different topological levels for constructing DReg loss on MobileNetV2 and ResNet18, respectively. \textbf{The results show that with a given training setup, DReg is more effective while DReg loss is introduced on the layer that closes to the output.} While for MobileNetV2, we observe that all tested topological levels outperform the vanilla momentum SGD training. However, for the more complex and robust model ResNet18, as the DReg loss moves closer to the input, DReg could converge to a worse point than vanilla momentum SGD. The reason is that DReg loss becomes the dominant portion of the overall loss, where SGD does not effectively optimize the cross-entropy loss.

    \pdfoutput=1
    
\begin{figure}[h!]
       \centering
    \begin{subfigure}{.4\textwidth}
      \centering
      \includegraphics[width=1.0\linewidth]{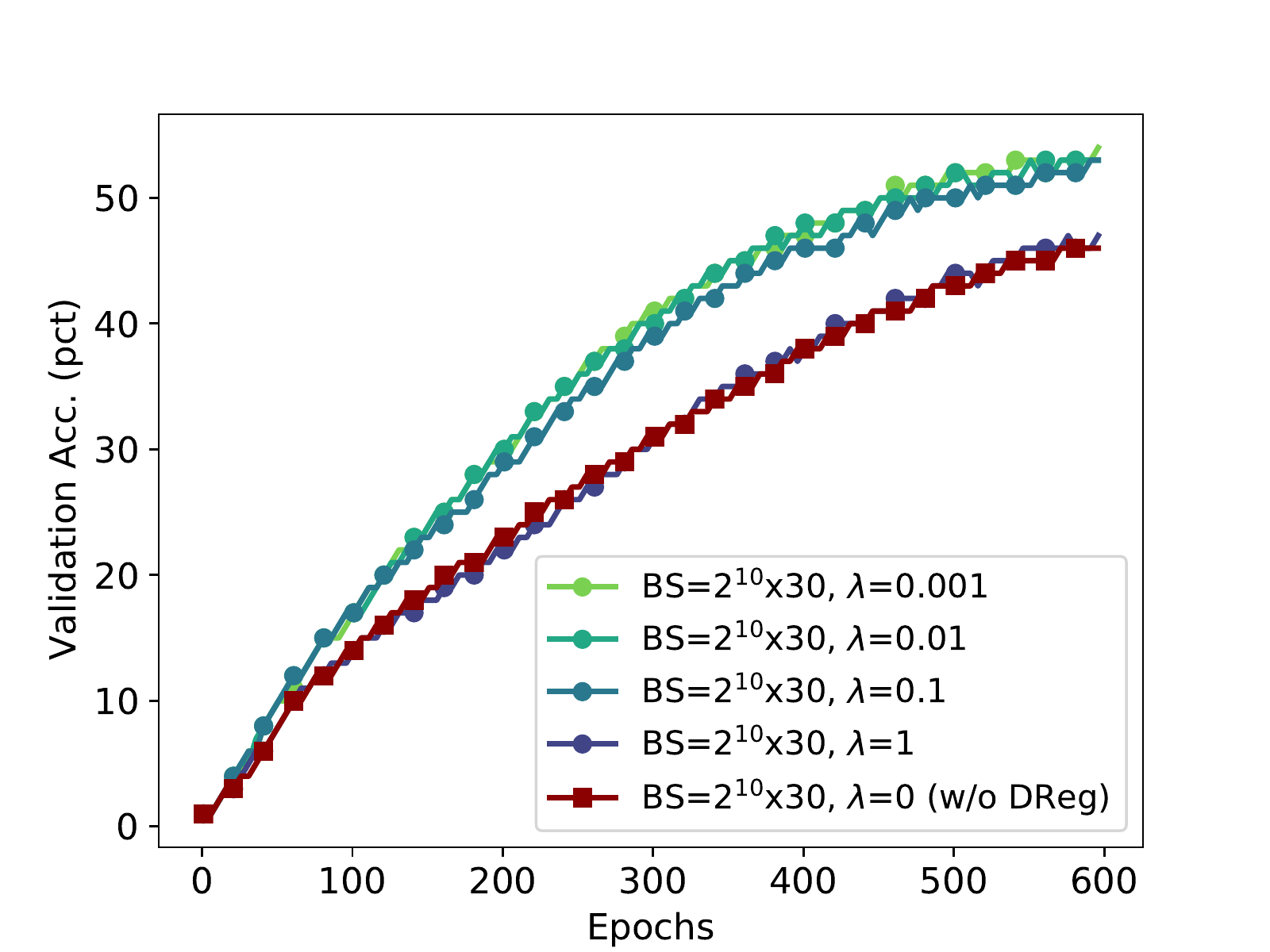}
      \caption{MobileNetV2, BS=$2^{10}\cdot30$}
      \label{fig:lambda_comparison_1}
    \end{subfigure}
        \begin{subfigure}{.4\textwidth}
      \centering
      \includegraphics[width=1.0\linewidth]{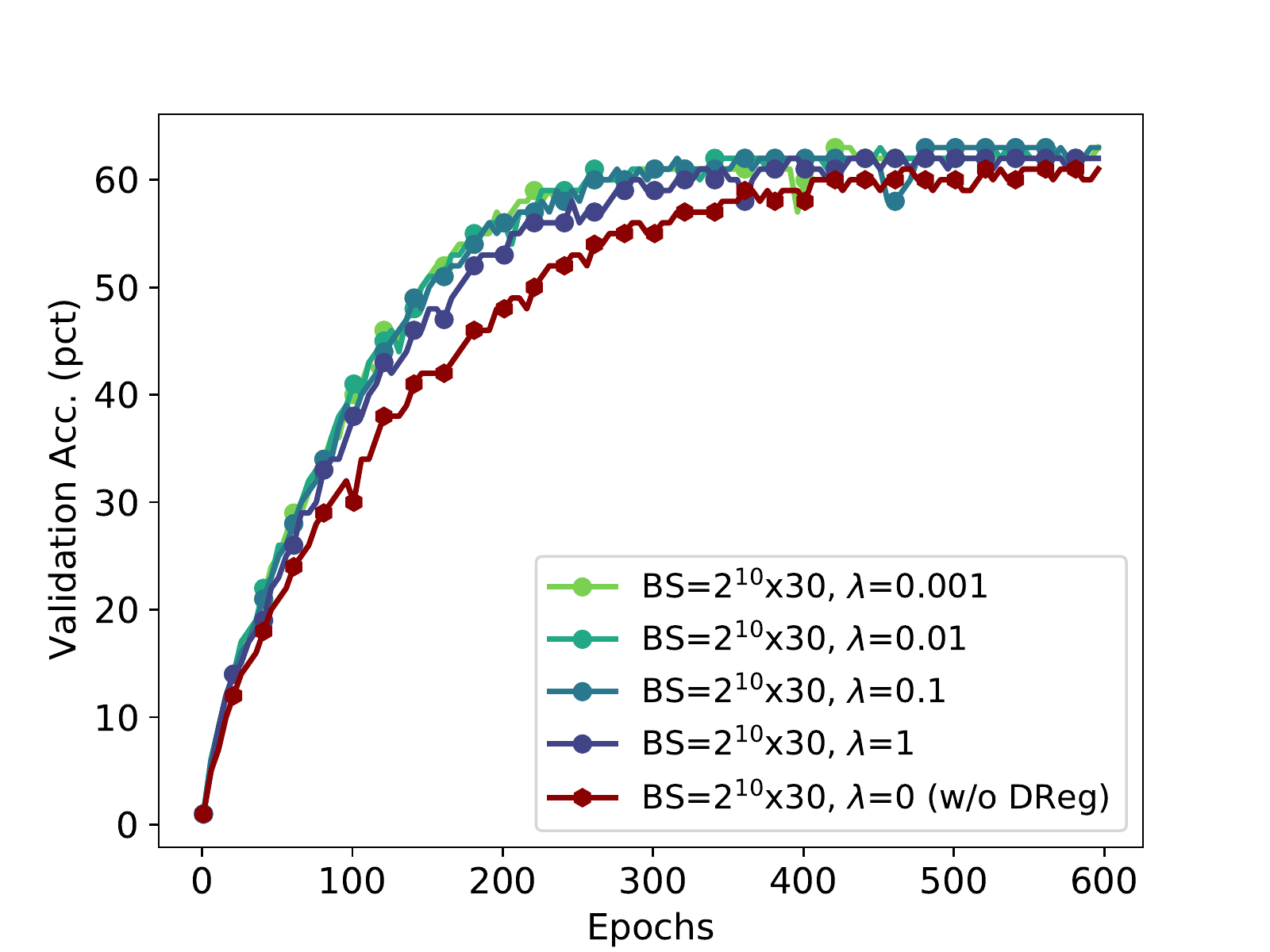}
      \caption{ResNet18, BS=$2^{10}\cdot30$}
      \label{fig:lambda_comparison_2}
    \end{subfigure}
    \caption{Evaluation of choosing different DReg loss weights $\lambda \in \{0.001, 0.01, 0.1, 1, 0\}$, using MobileNetV2 and ResNet18 on CIFAR100 dataset with momentum SGD.}
    \label{fig:lambda_comparison}
    \end{figure}

\textbf{Choices of $\lambda$} We evaluate the performance of applying different $\lambda$ values on MobileNetV2 and ResNet18 with last DW block used in DReg loss, shown in Figure \ref{fig:lambda_comparison}, with MobileNetV2 results in Figure \ref{fig:lambda_comparison_1} and ResNet18 results in Figure \ref{fig:lambda_comparison_2}. It turns out that the training performance is less sensitive to the choices of $\lambda$ compared to DReg topological level, since all DReg results in Figure \ref{fig:lambda_comparison} outperform vanilla momentum SGD. In addition, we observe small $\lambda$ values lead to slight better performance.

    \pdfoutput=1

\subsection{The DReg Impacts on Momentum in Minibatch SGD}\label{sec:momentum}

\begin{figure}[h!]
    \centering
    \begin{subfigure}{.48\textwidth}
      \centering
      \includegraphics[width=1.0\linewidth]{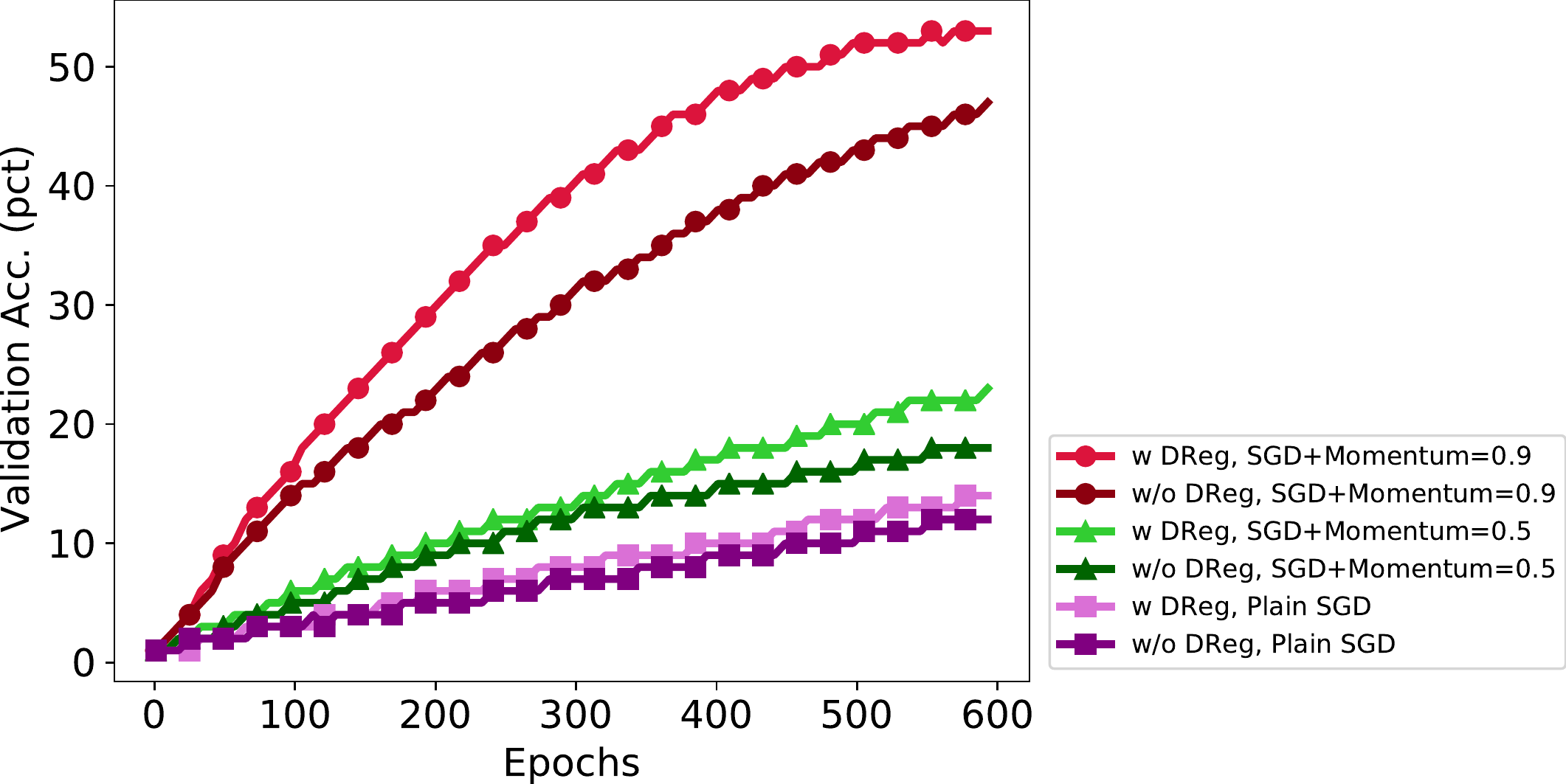}
      \caption{MobileNetV2, BS=$2^{10}\cdot30$}
      \label{fig:optimizers_acc_mobilenet_1}
    \end{subfigure}%
    \hspace{3mm}
        \begin{subfigure}{.48\textwidth}
      \centering
      \includegraphics[width=1.0\linewidth]{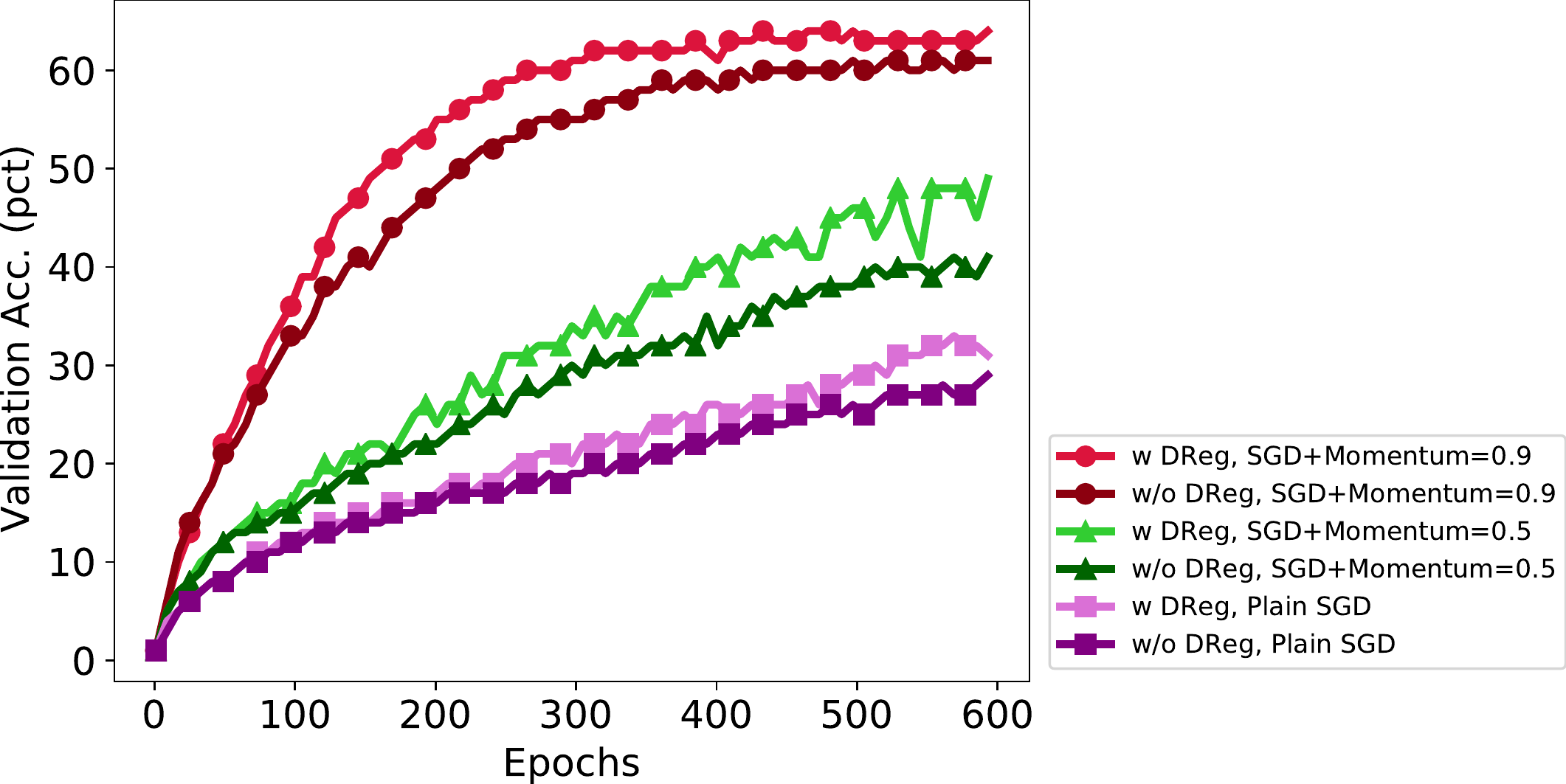}
      \caption{ResNet18, BS=$2^{10}\cdot30$}
      \label{fig:optimizers_acc_resnet_3}
    \end{subfigure}
    \caption{Comparisons of DReg in large minibatch training using momentum and vanilla SGD on CIFAR100 dataset, where $\lambda=0.01$ --- \textbf{(a)} MobileNetV2, BS$=2^{10}\cdot30$; \textbf{(b)} ResNet18, BS$=2^{10}\cdot30$.}
    \label{fig:optimizers_acc}
    \end{figure}

Finally, we demonstrate that DReg can accelerate both the vanilla and momentum minibatch SGD. The results shown in Figure \ref{fig:optimizers_acc} include optimizer setups (1) vanilla SGD, (2) SGD with momentum factor 0.9, and (3) SGD with momentum factor 0.5. We observe that DReg improves the minibatch SGD over large batch sizes regardless of the value of the momentum. Moreover, we observe that DReg can further boost the minibatch SGD if a larger momentum is applied, which shows that DReg also offers a momentum boost. For both experiments in Figure \ref{fig:optimizers_acc}, the generalization gap between vanilla SGD and distinctive-regularized SGD increases as the momentum factor increases.

\pdfoutput=1

\section{Conclusion}

In this paper, we present a novel regularization technique, distinctive regularization (DReg), that addresses the generalization gap of large minibatch SGD for training complex ConvNet models. The proposed DReg algorithm is orthogonal to the existing large minibatch SGD training techniques using learning rate scaling and gradient noises.  Our comprehensive experiments demonstrate a superior training boost for a wide range of batch sizes for training modern ConvNet models. We provide an empirical analysis of the DReg configurations to provide intuitions of constructing DReg for minibatch SGD. More importantly, our experimental results demonstrate that momentum in minibatch SGD is also benefited by using DReg.

\small
\bibliographystyle{plainnat}

\end{document}